\documentclass[pdflatex,sn-basic]{sn-jnl}% Basic Springer Nature Reference Style/Chemistry Reference Style

\usepackage{hyperref}       % hyperlinks
\usepackage{url}            % simple URL typesetting
\usepackage{graphicx}%
\usepackage{multicol}
\usepackage{multirow}%
\usepackage{makecell}%
\usepackage{amsmath,amssymb,amsfonts}%
\usepackage{amsthm}%
\usepackage{mathrsfs}%
\usepackage[title]{appendix}%
\usepackage{xcolor}%
\usepackage{textcomp}%
\usepackage{manyfoot}%
\usepackage{booktabs}%
\usepackage{algorithm}%
\usepackage{algorithmic}
\usepackage{listings}%

\usepackage{nicefrac}       % compact symbols for 1/2, etc.
\usepackage{microtype}      % microtypography
\usepackage{doi}
\usepackage{mathtools}
\usepackage{subfigure}
\usepackage{tabularx}
\usepackage{adjustbox}

\usepackage{hhline}
\usepackage{wrapfig}
\usepackage{siunitx}
\usepackage{enumitem}
\usepackage{dcolumn}
\usepackage{array}

\usepackage[authoryear]{natbib}
\setcitestyle{aysep={}}
\begin{document}

\title[Neural Architecture Search: Two Constant Shared Weights Initialisations]{Neural Architecture Search: Two Constant Shared Weights Initialisations}
%%=============================================================%%
%% GivenName	-> \fnm{Joergen W.}
%% Particle	-> \spfx{van der} -> surname prefix
%% FamilyName	-> \sur{Ploeg}
%% Suffix	-> \sfx{IV}
%% \author*[1,2]{\fnm{Joergen W.} \spfx{van der} \sur{Ploeg} 
%%  \sfx{IV}}\email{iauthor@gmail.com}
%%=============================================================%%

\author*[1]{\fnm{Ekaterina} \sur{Gracheva}}\email{gracheva.ekaterina@nims.go.jp}

\affil*[1]{\orgdiv{Center for Basic Research on Materials}, \orgname{National Institute for Materials Science}, \orgaddress{\street{1-1 Namiki}, \city{Tsukuba}, \postcode{3050044}, \state{Ibaraki}, \country{Japan}}}

%%==================================%%
%% Sample for unstructured abstract %%
%%==================================%%

\abstract{In the last decade, zero-cost metrics have gained prominence in neural architecture search (NAS) due to their ability to evaluate architectures without training. These metrics are significantly faster and less computationally expensive than traditional NAS methods and provide insights into neural architectures’ internal workings. This paper introduces \texttt{epsinas}, a novel zero-cost NAS metric that assesses architecture potential using two constant shared weight initialisations and the statistics of their outputs. We show that the dispersion of raw outputs, normalised by their average magnitude, strongly correlates with trained accuracy. This effect holds across image classification and language tasks on NAS-Bench-101, NAS-Bench-201, and NAS-Bench-NLP. Our method requires no data labels, operates on a single minibatch, and eliminates the need for gradient computation, making it independent of training hyperparameters, loss metrics, and human annotations. It evaluates a network in a fraction of a GPU second and integrates seamlessly into existing NAS frameworks. The code supporting this study can be found on GitHub at \url{https://github.com/egracheva/epsinas}.}


%%\pacs[JEL Classification]{D8, H51}

%%\pacs[MSC Classification]{35A01, 65L10, 65L12, 65L20, 65L70}

\maketitle

\section{Introduction}\label{sec1}

The field of neural architecture search (NAS) emerged roughly a decade ago as a solution to automate the process of neural geometry optimisation. In its early stages, NAS predominantly relied on evaluating candidate architectures through computationally intensive training processes, using techniques such as reinforcement learning \citep{williams1992simple}, evolutionary algorithms \citep{real2019regularized, liu2021survey}, and Bayesian optimisation \citep{falkner2018bohb, white2021bananas}.

One-shot algorithms speed up the search by sharing weights instead of training each architecture separately. These methods include efficient reinforcement learning \citep{pham2018efficient}, random search with parameter sharing \citep{li2020random}, and differentiable approaches \citep{liu2018darts, chu2020darts,xiang2023zero}. However, they require training a large hypernetwork, necessitating intricate hyperparameter tuning. While these techniques are efficient, they often fail to deliver consistent results \citep{dong2019searching}. For example, even advanced approaches like DARTS- \citep{chu2020darts} exhibit significant variability compared to evolutionary or reinforcement learning methods.

Some methods attempt to estimate network performance without training on the dataset of interest by leveraging an auxiliary predictive machine learning (ML) model built on a dataset of pre-evaluated architectures \citep{istrate2019tapas, deng2017peephole}. Although these methods accelerate the NAS process for image recognition, they still rely on training and are not applicable to other domains of ML.

Evaluating architectures through training has several disadvantages. The most obvious is computational expense, which makes large-scale evaluations infeasible for massive datasets. As a result, architectures are typically trained with a single random seed and a fixed set of hyperparameters, raising concerns about statistical reliability. The selected architecture may perform well only within the constraints of these hyperparameters, leading to potentially suboptimal choices. Training also implies using hand-labelled data, which introduces human error --- for instance, the ImageNet dataset is known to have a label error of about $6$\,$\rm\%$  \citep{northcutt2021pervasive}. Furthermore, most NAS methods provide limited insights into why a particular architecture is selected, leaving gaps in understanding the underlying principles.

\section{Related work}\label{related}
To simplify architecture search, zero-cost NAS methods identify optimal architectures without requiring full training. These methods leverage zero-cost proxies \citep{gaier2019weight, mellor2020neural, fan2023data, gracheva2021trainless, li2023zico}, typically demanding computational resources equivalent to just one or a few training epochs. Consequently, zero-cost NAS is two to three orders of magnitude faster than traditional NAS approaches. Below, we summarize the most prominent zero-shot NAS methods. For a comprehensive field review, see \cite{li2024zero}.

% \begin{unenumerate}
% \begin{description}[style=unboxed,leftmargin=0cm]
\textbf{Weight agnostic neural networks.}
% \item[Weight agnostic neural networks.]
One of the pioneering works in zero-shot NAS is presented by \cite{gaier2019weight}. The authors demonstrate a method for constructing neural architectures based on the mean accuracy over several initialisations with constant shared weights and the number of parameters within the model. The resulting model achieves over $90$\,$\rm\%$ accuracy on the MNIST dataset \citep{lecun2010mnist} when the weights are fixed to the best-performing constants. While these results are intriguing, the authors acknowledge that such architectures do not perform particularly well once trained. Moreover, in 2019, the benchmark databases of trained architectures, now routinely used to compare NAS metrics, had not yet been released, preventing a direct comparison of this zero-shot method with more recent approaches.

\textbf{ReLU activation pattern.}
% \item[ReLU activation pattern.]
In 2020, \cite{mellor2020neural} introduced the \texttt{naswot} metric, which exploits the property of the rectified linear unit (ReLU) activation function \citep{agarap2018deep} to produce distinct activation patterns for different architectures. Specifically, every image passing through a network generates a binary activation vector, which forms a binary matrix for a mini-batch. The logarithm of the determinant of this matrix serves as a scoring metric. The authors demonstrated that higher \texttt{naswot} values are associated with better training performance, concluding that high-performing networks should be able to distinguish inputs even before training. However, this method is limited to networks using ReLU activation functions, which restricts its applicability to convolutional architectures.

A conceptually similar work by \cite{chen2021neural} combines the number of linear regions in the input space with the spectrum of the neural tangent kernel (NTK) to develop the \texttt{tenas} metric. Instead of individually evaluating each network in the search space, they construct a super-network that includes all the available edges and operators, which is then pruned.

\textbf{Coefficient of variance.}
Our previous work on fully trainless NAS \citep{gracheva2021trainless} evaluated the stability of untrained scores across random weight initialisations. Networks were initialised with multiple random seeds, and architectures were selected based on the coefficient of variance of the accuracy at initialisation, \texttt{cv}. While \texttt{cv} performance is associated with a high error rate, the study concluded that a good architecture should demonstrate stability against random weight fluctuations. Although this method theoretically applies to any type of neural architecture, it requires multiple initialisations and is relatively computationally expensive compared to \texttt{naswot} and later methods. Additionally, accuracy-based scoring metrics are limited to classification problems, and it remains unclear how to adapt the \texttt{cv} approach for regression tasks.

\textbf{Inverse variance of gradients.}
A recent study by \cite{li2023zico} introduced the \texttt{zico} metric, a zero-cost proxy for neural architecture performance that, like \texttt{cv} \citep{gracheva2021trainless}, is derived from network statistics. \texttt{zico} is computed as the ratio of the average gradient values to their standard deviation—in essence, a reversed \texttt{cv}, where accuracy is substituted with gradient values. Notably, \texttt{zico} exhibits a strong correlation with trained accuracy, surpassing \texttt{nparams}—the total number of parameters in a network—as a performance estimator. While \texttt{nparams} may seem like a simplistic proxy, it remained the most effective predictor of network performance until 2023. 

\textbf{Gradient signal-to-noise ratio.}
In 2023, \cite{sun2023unleashing} introduced the gradient signal-to-noise ratio metric ($\xi$\texttt{-gsnr}), which, as the name suggests, closely resembles \texttt{zico}. Their approach builds on \texttt{zico} by squaring the metric and incorporating a small constant 
$\xi$ in the denominator. This seemingly minor modification significantly impacts the metric's performance, establishing a new state-of-the-art zero-cost NAS. Additionally, the authors provide essential theoretical justification for their approach. A comparison of the \texttt{cv}, \texttt{zico}, and 
$\xi$\texttt{-gsnr} methodologies is presented in Table \ref{tab:metrics_comparison}.

\textbf{Gradient sign.}
The \texttt{grad\_sign} metric is designed to approximate the sample-wise optimisation landscape \citep{zhang2021gradsign}. The authors argue that when local minima for various samples are closer, the probability increases that the corresponding gradients will have the same sign. The number of samples yielding the same gradient sign approximates this probability, allowing for the smoothness evaluation of the optimisation landscape and the architecture's trainability. However, this method requires labels and gradient computation.

\textbf{Pruning-at-initialisation proxies.}
% \item[Pruning-at-initialisation proxies.]
Several promising zero-cost proxies have emerged as adaptations of pruning-at-initialisation methods for NAS, as demonstrated in the work by \cite{abdelfattah2021zero}: \texttt{grad\_norm} \citep{wang2020picking}, \texttt{snip} \citep{lee2018snip}, and \texttt{synflow} \citep{tanaka2020pruning}. These metrics were originally developed to assess the salience of a network's parameters and prune potentially irrelevant synapses. They require a single forward-backwards pass to compute the loss, after which the importance of parameters is calculated as the product of the weight and gradient values. The salience is then integrated over all parameters in the network to estimate its potential performance once trained.

What makes the \texttt{synflow} metric particularly notable is its ability to evaluate architectures without referencing the data, as it computes the loss based solely on the product of all randomly initialised weights' values. Among these proxies, the \texttt{synflow} metric demonstrates the most consistent performance across various search spaces and sets the state-of-the-art for zero-cost NAS.
% \end{description}
% \end{unenumerate}

Neither \texttt{naswot} nor \texttt{synflow} depend on labels, which reduces the effect of human error during data labelling. Moreover, \texttt{naswot} does not require gradient computation, which renders this method less memory-intensive.

A few other efficient zero-shot methods cannot be directly compared to our results (such as Zen-NAS by \cite{lin2021zen}). For more details on the current status of NAS, please refer to the most recent systematic review on NAS \citep{salmani2025systematic}.

The results of the above studies suggest that neural networks possess an intrinsic property that determines their predictive potential even before training. This property should be independent of the specific values of trainable parameters (weights) and instead depend solely on the network's topology. In this work, we build upon insights from existing zero-cost NAS implementations to propose a new metric that shows strong performance and outperforms current zero-cost NAS methods.

\section{Epsinas metric}
Two existing NAS methods inspire the metric that we share in the present work: \texttt{cv} \citep{gracheva2021trainless} and weight agnostic neural networks \citep{gaier2019weight}. Both metrics aim to reduce the role of individual weights in the network to access its topology. The \texttt{cv} metric levels out the individual weights via multiple random initialisations, while \cite{gaier2019weight} sets the weights to the same value across the network. 

As mentioned, the \texttt{cv} metric has two principal disadvantages. While it shows a fairly consistent trend with trained accuracy, it suffers from high statistical deviations. To some degree, it must be due to the noise from random weight initialisations. With \texttt{epsinas}, we replaced random initialisations with single shared weight initialisations to improve the method's performance.

The second weak point of \texttt{cv} is that it has been developed for classification problems and relies on accuracy, which should be maximised. On the other hand, for regression tasks, performance is typically computed as an error, which is sought to be minimised. The \texttt{cv} is a ratio of standard deviation and the mean of untrained accuracies, and the final metric correlates negatively with train accuracy. It is not apparent whether the division by mean untrained \textit{error} would result in the same trend for \texttt{cv} metric. To address this issue, we decided to step out from evaluation metrics and consider raw outputs instead. This modification renders the method applicable to any neural architecture. 

For this, we flatten the raw output matrices to obtain a single vector $\mathbf{v}$ of length $L_\mathbf{v}=N_{\textrm{BS}} \times L_\mathbf{O}$ per initialisation, where $N_{\rm{BS}}$ is the batch size and $L_\mathbf{O}$ is the length of a single output \footnote{This length depends on the task and architecture: for regression tasks $L_\mathbf{O}=1$, for classification, $L_\mathbf{O}$ is equal to the number of classes, and for recurrent networks, it depends on the desired length of generated string.}. 

%We then stuck both initialisations into a single output matrix $\mathbf{V}$. 

Before proceeding to statistics computation over initialisations, we also must normalise the output vectors: in the case of constant shared weights, outputs scale with weight values. To compare initialisations on par with each other, we use min-max normalisation:

\begin{equation}
\mathbf{v}_{i}^\prime=\frac{\mathbf{v}_{i}-\min(\mathbf{v}_{i})}{\max(\mathbf{v}_{i})-\min(\mathbf{v}_{i})},\
\label{eqn:minmax}
\end{equation}

where $i$ is the index for initialisations, $i=\{0,1\}$.

We noticed that two distinct weights are sufficient to grasp the difference between initialisations. Accordingly, instead of standard deviation, we use mean absolute error between the normalised outputs of two initialisations:

\begin{equation}
\mathrm{MAE} = \frac{1}{L_\mathbf{v}} \sum\limits_{j=0}^{L_\mathbf{v}}\lvert \mathbf{v}_{0,j}^\prime - \mathbf{v}_{1,j}^\prime\rvert.
\label{eqn:delta}
\end{equation}

The mean is computed over the outputs of both initialisations as follows:

\begin{equation}
\mu = \frac{1}{L_\mathbf{v}} \sum\limits_{j=0}^{L_\mathbf{v}} \frac{\mathbf{v}_{1,j}^\prime + \mathbf{v}_{2,j}^\prime}{2}
\label{eqn:mean}
\end{equation}

Finally, the metric is computed as the ratio of $\mathrm{MAE}$ and $\mu$:

\begin{equation}
\varepsilon = \frac{\mathrm{MAE}}{\mu}.
\label{eqn:epsinas}
\end{equation}

We use real data from the respective datasets to compute the statistics, as suggested in \cite{fan2023data}. In Section \ref{sec:synthetic_data}, we demonstrate that using real data enhances search efficiency.

We refer to our metric as \texttt{epsinas}, paying tribute to the $\varepsilon$ symbol commonly used in mathematics to denote error bounds. The \texttt{epsinas} metric was developed independently but in parallel with \texttt{zico} and follows a similar approach. While \texttt{zico} uses the mean and standard deviation of gradients over two mini-batches, our metric combines the mean and mean absolute error of raw network outputs. This parallel development highlights a shared idea — leveraging statistical properties of untrained networks to guide architecture selection.

Table \ref{tab:metrics_comparison} compares \texttt{epsinas} to the three the most resembling zero-cost proxies: \texttt{cv}, \texttt{zico}, and $\xi$\texttt{-gsnr}. Although the difference between the two methods may seem minor and \texttt{zico} and $\xi$\texttt{-gsnr} show a strong correlation with trained accuracy, \texttt{epsinas} has an advantage: it does not require backpropagation for gradients computation, making it faster to calculate. Additionally, as we demonstrate in the next section, \texttt{epsinas} exhibits an even stronger correlation with trained accuracy.

\begin{table*}[!t]
\caption{Comparison of \texttt{epsinas} to the three the most resembling zero-cost proxies. Data $\mathbf{D}$ denotes the data used for metrics computation. $N_{init}$ refers to the number of initialisations (for \texttt{zico} and $\xi$\texttt{-gsnr} we report the number of batches).}
\label{tab:metrics_comparison}
\begin{center}
\small
\renewcommand{\tabcolsep}{0.07cm}
\renewcommand{\arraystretch}{1.2}
\begin{tabular}{llp{2.0cm}llcc}
\toprule
\noalign{\vskip 1 pt} 
Metric & Data, $\mathbf{D}$ & Formula & Initialisation & $N_{init}$ & Gradients & True labels \\
\noalign{\vskip 1pt} 
\midrule
\noalign{\vskip 1pt} 
\texttt{cv}         & Accuracy           & $\frac{\sigma(\mathbf{D})}{\mu(\mathbf{D})}$         & \makecell[l]{Kaiming \\normal} & 100 & No  & Yes \\
\texttt{zico}       & Gradients          & $\frac{\mu(\mathbf{D}}{\sigma(\mathbf{D})}$          & \makecell[l]{Kaiming \\normal} & 2   & Yes & Yes \\
$\xi$\texttt{-gsnr} & Gradients          & $\frac{\mu(\mathbf{D})^2}{\sigma(\mathbf{D})^2+\xi}$ & \makecell[l]{Kaiming \\normal} & 8   & Yes & Yes \\
\texttt{epsinas}    & \makecell[l]{Normalised \\outputs} & $\frac{\mathrm{MAE}(\mathbf{D})}{\mu(\mathbf{D})}$           & \makecell[l]{Constant \\shared weights} & 2   & No  & No  \\
\bottomrule
\end{tabular}
\end{center}
\end{table*}

Algorithm \ref{algo} details the \texttt{epsinas} metric computation.

\begin{algorithm}[tb]
\caption{Algorithm for \texttt{epsinas} metric computation}
\label{algo}
\begin{algorithmic}
    \STATE Select a \texttt{batch} of data from train set
    \FOR{\texttt{arch} in \texttt{search space}}
        \STATE Initialise empty output \texttt{matrix}
        \FOR{\texttt{weight} in [\texttt{val1}, \texttt{val2}]}
            \STATE Initialise \texttt{arch} with constant shared \texttt{weight}
            \STATE Forward pass the \texttt{batch} through \texttt{arch}
            \STATE Get and flatten \texttt{outputs}
            \STATE Minmax normalise \texttt{outputs} (Eq. \ref{eqn:minmax})
            \STATE Append \texttt{outputs} to the output \texttt{matrix}
        \ENDFOR
        \STATE Compute difference between the rows of the output \texttt{matrix} (Eq. \ref{eqn:delta})
        \STATE Compute mean over the output \texttt{matrix} (Eq. \ref{eqn:mean})
        \STATE Compute \texttt{epsinas} metric (Equation \ref{eqn:epsinas})
    \ENDFOR
\end{algorithmic}
\end{algorithm}

Proper initialization is crucial for implementing the metric effectively. In our current implementation, we preserve the initialization of biases, batch normalization, and embeddings. Specifically, bias weights are set to zero, batch normalization parameters are initialized to ones, and embeddings follow the original paper’s approach using random integers \citep{klyuchnikov2022bench}.

\section{Experimental design}

\subsection{NAS benchmarks}
Following standard NAS metric evaluation practices, we assess \texttt{epsinas} on three well-established NAS benchmark datasets containing fully trained neural architectures. To ensure the broad applicability of \texttt{epsinas} across different architectures and tasks, we evaluate it on both image classification and language processing tasks.

\textbf{NAS-Bench-101.}
The first and one of the most extensive NAS benchmarks, NAS-Bench-101, consists of $423{,}624$ fully trained convolutional neural networks \citep{ying2019bench}. Each architecture comprises three stacked cells, followed by max-pooling layers, with up to $7$ vertices, $9$ edges, and $3$ possible operations per cell. All models are trained on \mbox{CIFAR-10} \citep{krizhevsky2009learning} for $108$ epochs using fixed hyperparameters.

\textbf{NAS-Bench-201.}
This benchmark provides $15{,}625$ architectures with a fixed backbone: a convolutional layer followed by three stacked cells connected by a residual block \citep{dong2020bench}. Each cell is a densely connected directed acyclic graph with $4$ nodes and $5$ possible operations. Architectures are trained on \mbox{CIFAR-10}, \mbox{CIFAR-100} \citep{krizhevsky2009learning}, and a downsampled version of ImageNet \citep{chrabaszcz2017downsampled}, all with fixed hyperparameters over $200$ epochs.

\textbf{NAS-Bench-NLP.}
Recurrent cells comprise $24$ nodes, $3$ hidden states and $3$ input vectors at most, with $7$ allowed operations. Here, we only consider models trained and evaluated on Penn Treebank (PTB) data \citep{marcinkiewicz1994building}: $14{,}322$ random networks with a single seed. The training spans $50$ epochs and is conducted with fixed hyperparameters.

\subsection{Evaluation metrics}
We evaluate the performance of \texttt{epsinas} and compare it to the results for zero-cost NAS metrics reported in \cite{abdelfattah2021zero}. We use the following evaluation scores (computed with \texttt{NaN} omitted):

\begin{itemize}
\itemsep-0.2em 
\item Spearman $\rho$ (global): Spearman rank correlation $\rho$ evaluated on the entire dataset. 
\item Spearman $\rho$ (top-$10\%$):
Spearman rank correlation $\rho$ for the top-$10\%$ performing architectures.
\item Kendall $\tau$ (global): Kendall rank correlation coefficient $\tau$ evaluated on the entire dataset.
\item Kendall $\tau$ (top-$10\%$): Kendall rank correlation coefficient $\tau$ for the top-$10\%$ performing architectures.
\item Top-$10\%$/top-$10\%$: fraction of top-$10\%$ performing models within the top-$10\%$ models ranked by zero-cost scoring metric ($\%$).
\item Top-64/top-$10\%$: number of top-64 models ranked by zero-cost scoring metric within top-$5\%$ performing models.
\end{itemize}

\section{Results}
\subsection{NAS-Bench-201}
The results for overall \texttt{epsinas} performance on NAS-Bench-201 are given in Table \ref{tab:results_201} along with other zero-cost NAS metrics. The Kendall $\tau$ score is not reported in \cite{abdelfattah2021zero}, but it is considered more robust than Spearman $\rho$ and is increasingly used for NAS metric evaluation. We use the data provided by \cite{abdelfattah2021zero} to evaluate their Kendal $\tau$. Note that our results differ from the original paper in terms of evaluation scores. In such cases, we indicate the original values between brackets. In particular, there is a discrepancy in computing the values in the last column, \textit{Top-64/top-$5\%$}, while the rest of the results are consistent. Figure \ref{fig:other_results_201} in the Appendix suggests that our calculations are correct.

% Table with rho and tau for NAS-Bench-201
\begin{table*}[t!]
\caption{Zero-cost metrics performance for the NAS-Bench-201 search space with its three datasets: CIFAR-10, CIFAR-100 and ImageNet16-120. We give the original values from \cite{abdelfattah2021zero} for reference between brackets. We highlight the best-performing metrics in bold (p-values between \texttt{epsinas} and other metrics are below 0.001).}
\label{tab:results_201}
\begin{center}
\footnotesize
\renewcommand{\tabcolsep}{0.07cm}
\renewcommand{\arraystretch}{1.2}
\begin{tabular}{lrrrrp{0.1mm}rrrrrr}
\toprule
\noalign{\vskip 1 pt} 
\multirow{2}{*}{Metric} & \multicolumn{4}{c}{Spearman $\rho$} & & \multicolumn{2}{c}{Kendall $\tau$} & \multicolumn{2}{c}{Top-10\%/} & \multicolumn{2}{c}{Top-64/} \\
% \cline{2-5}\cline{7-8}
\noalign{\vskip 1.5pt} 
  & \multicolumn{2}{c}{global} & \multicolumn{2}{c}{top-10\%} & & \multicolumn{1}{c}{global} & \multicolumn{1}{c}{top-10\%} & \multicolumn{2}{c}{top-10\%}& \multicolumn{2}{c}{top-5\%} \\
\midrule
\noalign{\vskip 1pt} 
\multicolumn{12}{c}{CIFAR-10}\\
\midrule
\noalign{\vskip 1pt} 
\texttt{grad\_sign} &  0.77  &  &  &  &  &  &  &  &  &  & \\
\texttt{synflow}    &  0.74  &        &  0.18  &         &  &  0.54  &  0.12 & 45.75 & (46) & 29 & (44) \\
\texttt{grad\_norm} &  0.59  & (0.58) & -0.36  & (-0.38) &  &  0.43  & -0.21 & 30.26 & (30) &  1 &  (0) \\
\texttt{grasp}      &  0.51  & (0.48) & -0.35  & (-0.37) &  &  0.36  & -0.21 & 30.77 & (30) &  3 &  (0) \\
\texttt{snip}       &  0.60  & (0.58) & -0.36  & (-0.38) &  &  0.44  & -0.21 & 30.65 & (31) &  1 &  (0) \\
\texttt{fisher}     &  0.36  &        & -0.38  &         &  &  0.26  & -0.24 &  4.99 & ( 5) &  0 &  (0) \\
\texttt{nparams}    &  0.75  &        &     &         &  &  0.57  &     &     &    &     &      \\
\texttt{zico}    &  0.80  &        &     &         &      &  0.61  &    &      &    &    &      \\
$\xi$\texttt{-gsnr}   &  0.845  &        &     &         &      &  0.661  &    &      &    &    &      \\
\texttt{epsinas}    &  \pmb{0.87}  &        &  \pmb{0.50}  &         &  &  \pmb{0.69}  &  \pmb{0.36} &  \pmb{67.06} &    & \pmb{62} &      \\
\midrule
\noalign{\vskip 1pt} 
\multicolumn{12}{c}{CIFAR-100}\\
\midrule
\texttt{grad\_sign} &  0.79  &  &  &  &  &  &  &  &  &  & \\
\texttt{synflow}    &  0.76  &        &  0.42  &         &  &  0.57  &  0.29 & 49.71 & (50) & 45 & (54) \\
\texttt{grad\_norm} &  0.64  &        & -0.09  &         &  &  0.47  & -0.05 & 35.00 & (35) &  0 &  (4) \\
\texttt{grasp}      &  0.55  & (0.54) & -0.10  & (-0.11) &  &  0.39  & -0.06 & 35.32 & (34) &  3 &  (4) \\
\texttt{snip}       &  0.64  & (0.63) & -0.08  & (-0.09) &  &  0.47  & -0.05 & 35.25 & (36) &  0 &  (4) \\
\texttt{fisher}     &  0.39  &        & -0.15  & (-0.16) &  &  0.28  & -0.10 &  4.22 & ( 4) &  0 &  (0) \\
\texttt{nparams}    &  0.73  &        &     &         &  &  0.55  &     &     &    &     &      \\
\texttt{zico}    &  0.81  &        &     &         &  &  0.61  &     &     &    &     &      \\
$\xi$\texttt{-gsnr}    &  0.840  &        &     &         &      &  0.658  &    &      &    &    &      \\
\texttt{epsinas}    &  \pmb{0.89}  &        &  \pmb{0.63}  &         &  & \pmb{0.72}  & \pmb{0.45} & \pmb{81.33} &    & \pmb{62} &      \\
\noalign{\vskip 1pt} 
\midrule

\multicolumn{12}{c}{ImageNet16-120}\\
\midrule

\texttt{grad\_sign}    &  0.78  &  &  &  &  &  &  &  &  &  & \\
\texttt{synflow}    &  0.75  &        & \pmb{0.55} &          &  &  0.56  & \pmb{0.39} & 43.57 & (44) & 26 & (56) \\
\texttt{grad\_norm} &  0.58  &        &  0.12 &  (0.13)  &  &  0.43  &  0.09 & 31.29 & (31) &  0 & (13) \\
\texttt{grasp}      &  0.55  & (0.56) &  0.10 &          &  &  0.39  &  0.07 & 31.61 & (32) &  2 & (14) \\
\texttt{snip}       &  0.58  &        &  0.13 &          &  &  0.43  &  0.09 & 31.16 & (31) &  0 & (13) \\
\texttt{fisher}     &  0.33  &        &  0.02 &          &  &  0.25  &  0.01 &  4.61 & ( 5) &  0 &  (0) \\
\texttt{nparams}    &  0.69  &        &     &         &  &  0.52  &     &     &    &     &      \\
\texttt{zico}    &  0.79  &        &     &         &  &  0.60  &     &     &    &     &      \\
$\xi$\texttt{-gsnr}    &  0.793  &        &     &         &      &  0.608  &    &      &    &    &      \\
\texttt{epsinas}    & \pmb{0.85}  &        &  0.53 &          &  & \pmb{0.67}  &  0.37 & \pmb{71.51} &    & \pmb{59} &   \\
\bottomrule
\end{tabular}
\end{center}
\end{table*}

For NAS-Bench-201, we also report average performance when selecting one architecture from a pool of $N$ random architectures. The statistics are reported over $500$ runs. Table \ref{tab:avrg_performance} compares \texttt{epsinas} to other zero-cost metrics. 

% Table, average performance
\begin{table*}[t!]
\caption{Comparison of the zero-cost metrics performances against existing NAS algorithms on CIFAR-10, CIFAR-100 and ImageNet16-120 datasets. On the top, we list the best-performing methods that require training: REA \citep{real2019regularized}, random search, REINFORCE \citep{williams1992simple}, BOHB \citep{falkner2018bohb}, DARTS- \citep{chu2020darts}. We report the average best-achieved test accuracy over $500$ runs, with $1{,}000$ architectures ($100$ for \texttt{grad\_sign}) sampled from the search space at random. Random and optimal performances are given as a baseline.}
\label{tab:avrg_performance}
\begin{center}
\footnotesize
\renewcommand{\tabcolsep}{0.07cm}
\renewcommand{\arraystretch}{1.2}
\begin{tabular}{llcccccccc}
\hline
\multirow{2}{*}{Method} & Cost & \multicolumn{2}{c}{CIFAR-10} & \multirow{2}{*}{} & \multicolumn{2}{c}{CIFAR-100} & \multirow{2}{*}{} & \multicolumn{2}{c}{ImageNet16-120}  \\
\cmidrule{3-4}\cmidrule{6-7}\cmidrule{9-10}
 & (h) & validation & test &   & validation & test &   & validation & test \\
\hline
\noalign{\vskip 1pt}
\multicolumn{1}{c}{} & \multicolumn{9}{c}{State-of-the-art}\\
\texttt{REA}             & $3.3$ & $91.19 \pm 0.31$ & $93.92 \pm 0.3$  & & $71.81 \pm 1.12$ & $71.84 \pm 0.99$ & & $45.15 \pm 0.89$  & $45.54 \pm 1.03$ \\
\texttt{RS}   & $3.3$ & $90.93 \pm 0.36$ & $93.92 \pm 0.31$ & & $70.93 \pm 1.09$ & $71.04 \pm 1.07$ & & $44.45 \pm 1.1$   & $44.57 \pm 1.25$ \\
\texttt{REINFORCE}       & $3.3$ & $91.09 \pm 0.37$ & $93.92 \pm 0.32$ & & $71.61 \pm 1.12$ & $71.71 \pm 1.09$ & & $45.05 \pm 1.02$  & $45.24 \pm 1.18$ \\
\texttt{BOHB}            & $3.3$ & $90.82 \pm 0.53$ & $93.92 \pm 0.33$ & & $70.74 \pm 1.29$ & $70.85 \pm 1.28$ & & $44.26 \pm 1.36$  & $44.42 \pm 1.49$ \\
\texttt{DARTS-}          & $3.2$ & $91.03 \pm 0.44$ & $93.80 \pm 0.40$ & & $71.36 \pm 1.51$ & $71.53 \pm 1.51$ & & $44.87 \pm 1.46$  & $45.12 \pm 0.82$ \\ 
\hline
\noalign{\vskip 1pt} 
\multicolumn{1}{c}{} & \multicolumn{9}{c}{Baselines (N=1000)}\\
\texttt{Optimal} & n/a & $91.34 \pm 0.18$ & $94.20 \pm 0.13$ & & $72.53 \pm 0.53$ & $72.84 \pm 0.41$ & & $45.93 \pm 0.51$  & $46.59 \pm 0.34$  \\
\texttt{Random}  & n/a & $84.11 \pm 11.71$ & $87.40 \pm 11.94$ & & $61.57 \pm 11.35$ & $61.67 \pm 11.35$ & & $33.97 \pm 8.68$  & $33.67 \pm 8.98$  \\
\hline
\noalign{\vskip 1pt} 
\multicolumn{1}{c}{} & \multicolumn{9}{c}{Zero-cost (N=1000)}\\
\texttt{naswot}  & $3.3$ & $89.69 \pm 0.73$ & $92.96 \pm 0.81$ &  & $69.86 \pm 1.21$ & $69.98 \pm 1.22$ & & $43.95 \pm 2.05$  & $44.44 \pm 2.10$ \\
\texttt{synflow}  & n/a & $89.91 \pm 0.83$ & $90.12 \pm 0.78$ &  & $70.35 \pm 2.25$ & $70.37 \pm 2.08$ & & $41.73 \pm 3.91$  & $42.11 \pm 4.02$ \\
\texttt{grad\_norm} & n/a & $88.13 \pm 2.35$ & $88.42 \pm 2.28$ &  & $66.35 \pm 5.45$ & $66.48 \pm 5.32$ & & $33.88 \pm 11.46$ & $33.90 \pm 11.74$ \\
\texttt{grasp}  & n/a & $87.85 \pm 2.12$ & $88.17 \pm 2.04$ &  & $65.36 \pm 5.57$ & $65.45 \pm 5.48$ & & $32.23 \pm 10.95$ & $32.20 \pm 11.23$ \\
\texttt{snip}   & n/a & $87.47 \pm 2.19$ & $87.81 \pm 2.12$ &  & $64.61 \pm 5.52$ & $64.74 \pm 5.43$ & & $30.65 \pm 11.32$ & $30.55 \pm 11.55$ \\
\texttt{fisher} & n/a & $87.01 \pm 2.31$ & $87.36 \pm 2.23$ &  & $63.54 \pm 5.69$ & $63.67 \pm 5.62$ & & $26.70 \pm 10.83$ & $29.56 \pm 10.83$ \\
\texttt{epsinas} & $0.03$ & $\pmb{91.03} \pm \pmb{0.42}$ & $\pmb{93.86} \pm \pmb{0.43}$ & & $\pmb{71.76} \pm \pmb{0.90}$ & $\pmb{71.79} \pm \pmb{0.86}$ & & $\pmb{45.11} \pm \pmb{0.99}$  & $\pmb{45.42} \pm \pmb{1.21}$ \\
\hline
\noalign{\vskip 1pt} 
\multicolumn{1}{c}{} & \multicolumn{9}{c}{Zero-cost (N=100)}\\
\texttt{cv}  & $0.05$ & $84.89 \pm 6.39$ & $91.90 \pm 2.27$ & & $63.99 \pm 5.61$ & $64.08\pm 5.63$ & & $38.68 \pm 6.34$ & $38.76 \pm 6.62$ \\
\texttt{grad\_sign} & n/a & $89.84 \pm 0.61$ & $93.31 \pm 0.47$ & & $70.22 \pm 1.32$ & $70.33 \pm 1.28$ & & $42.07 \pm 2.78$  & $42.42 \pm 2.81$ \\
\texttt{epsinas}  & $0.003$ & $90.44 \pm 0.97$ & $93.39 \pm 0.82$ & & $70.85 \pm 1.30$ & $71.00 \pm 1.26$ & & $44.03 \pm 2.02$  & $44.20 \pm 2.04$ \\
\hline
\end{tabular}
\end{center}
\end{table*}

Comparing the results for \texttt{epsinas} with other zero-cost NAS metrics, we can see that it shows surprisingly good performance given its conceptual simplicity. Importantly, it is consistently better than the naive estimator \texttt{nparams}.

\begin{figure*}[!t]
\centering
\resizebox{\textwidth}{!}{%
\includegraphics[height=3cm]{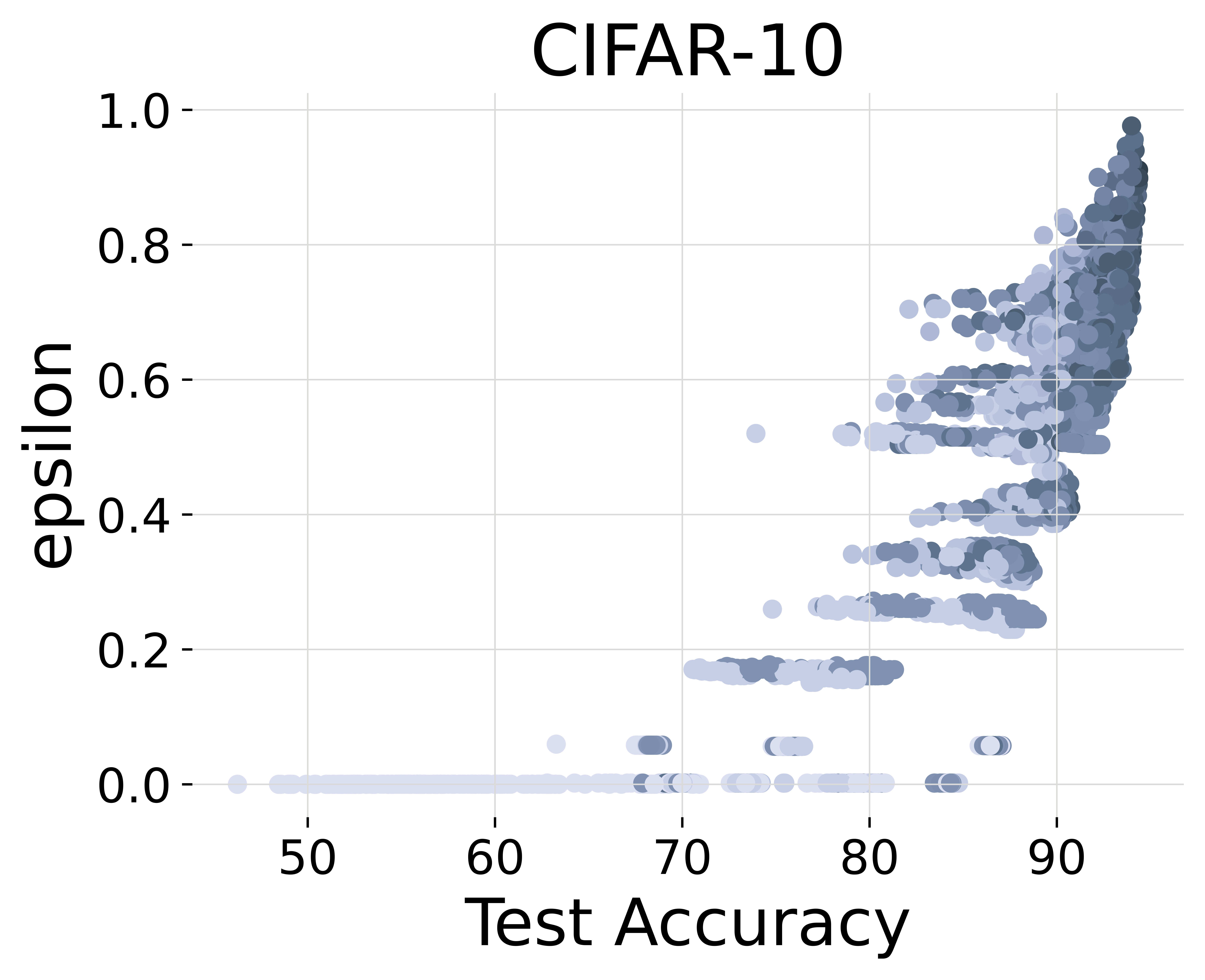}\hfill
\includegraphics[height=3cm]{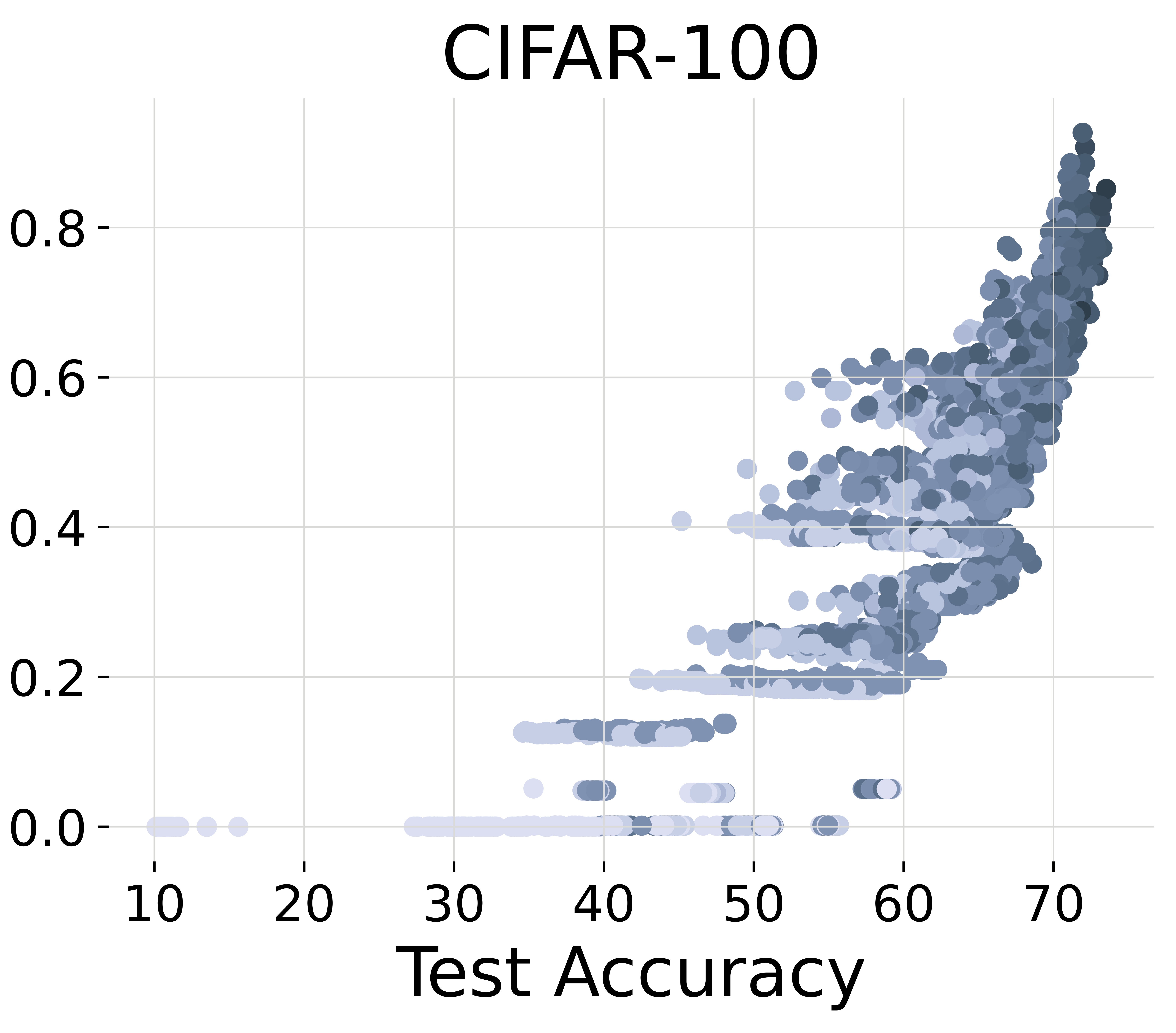}\hfill
\includegraphics[height=3cm]{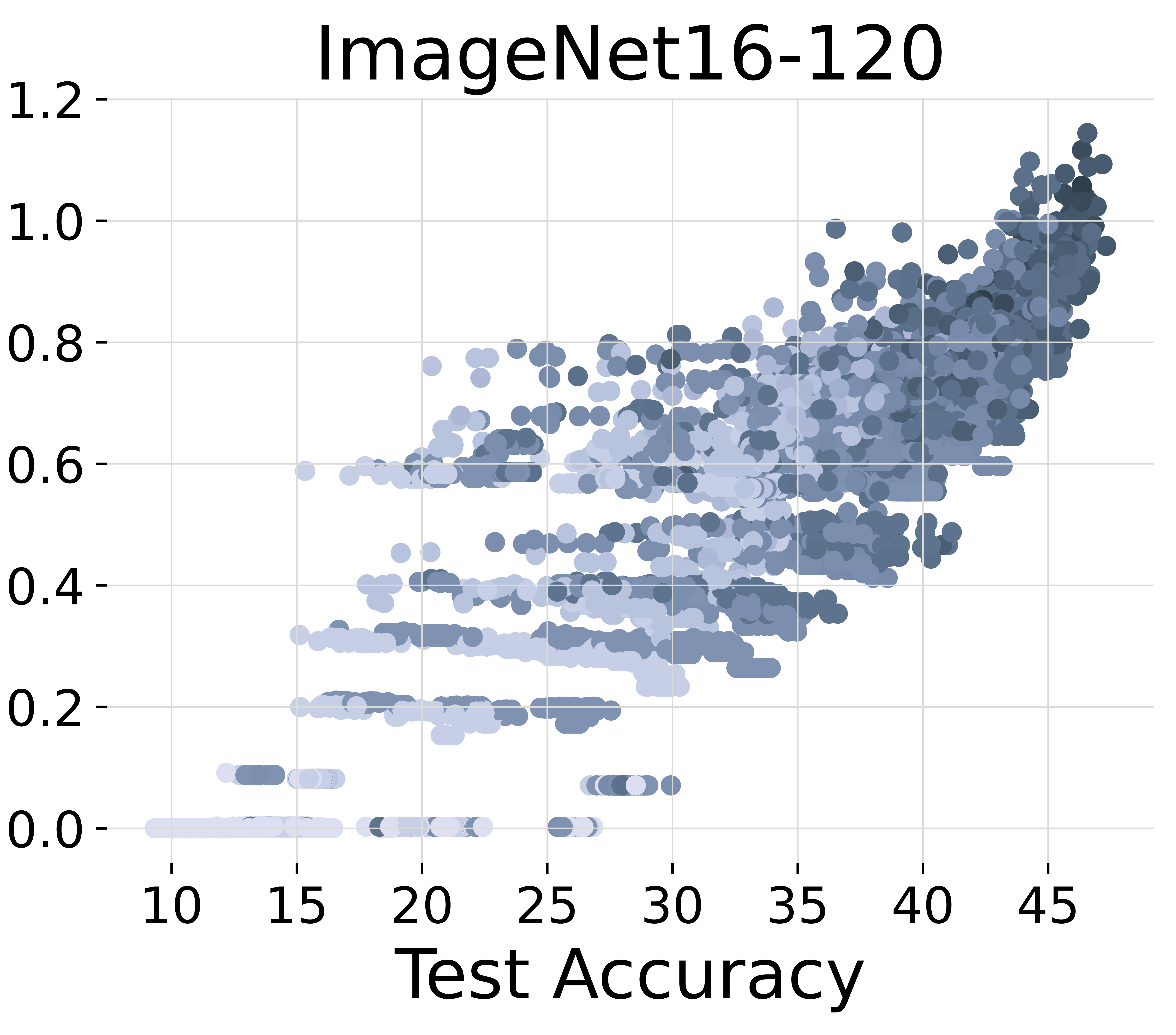}
}
\caption{Zero-cost NAS \texttt{epsinas} metric performance illustration for NAS-Bench-201 search space evaluated on CIFAR-10, CIFAR-100 and ImageNet16-120 datasets. The horizontal axis shows test accuracy upon training. Each dot corresponds to an architecture; the darker the colour, the more
parameters it contains. The figure represents the search space of $15{,}625$ networks (excluding architectures with \texttt{NaN} scores).}
\label{fig:results_201}
\end{figure*}

Figure \ref{fig:results_201} further confirms the applicability of the method to the NAS-Bench-201 field (similar plots for other methods can be found in the Ablation section, Figure \ref{fig:other_results_201}). However, NAS-Bench-201 is a relatively compact search space; furthermore, it has been used for \texttt{epsinas} development.

% Figure, epsinas vs Accuracy, NAS-Bench-201

\subsection{NAS-Bench-101}
\begin{figure*}[!t]
\begin{center}
\resizebox{0.7\textwidth}{!} {%
\includegraphics[width=2.5in]{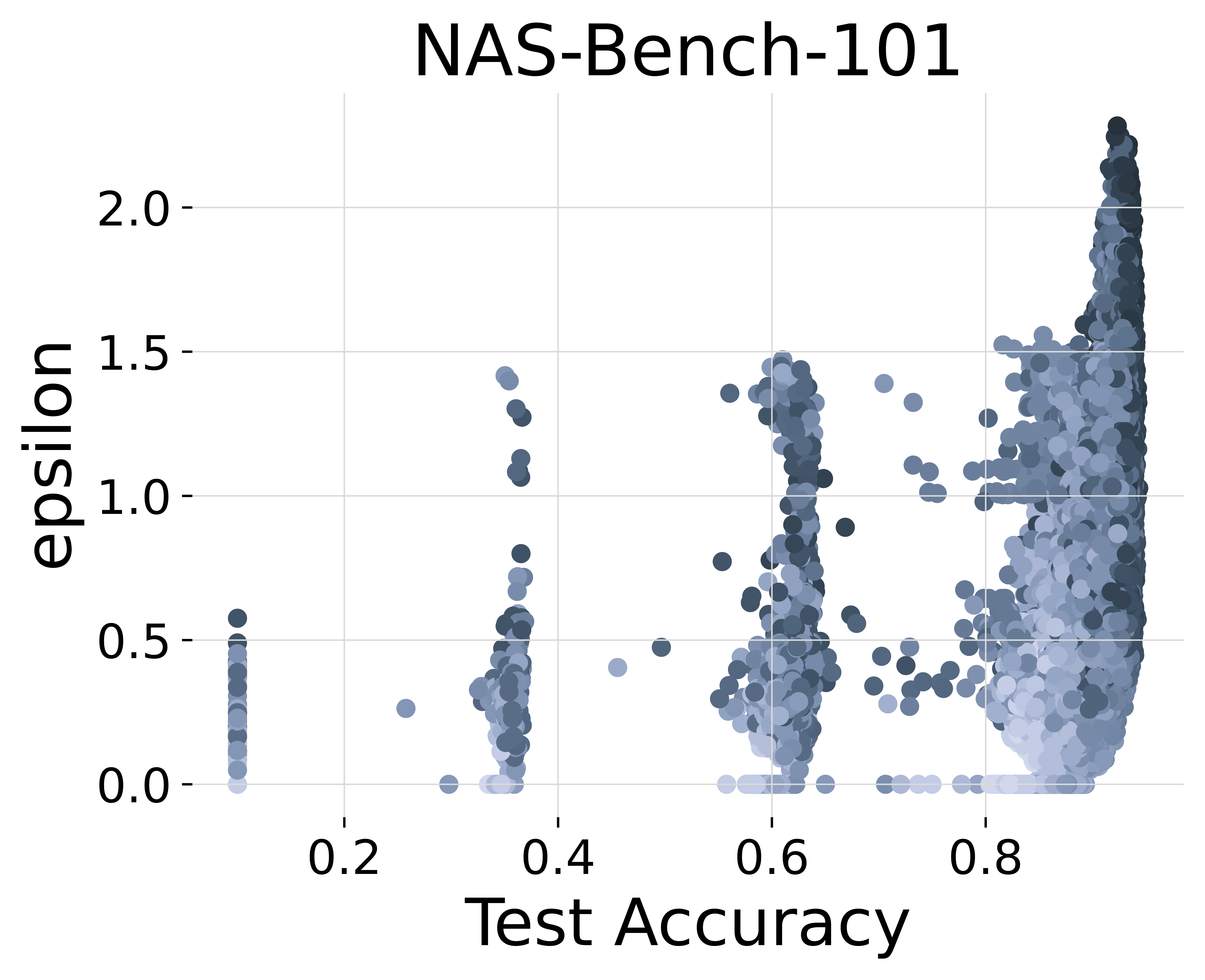}
\includegraphics[width=2.5in]{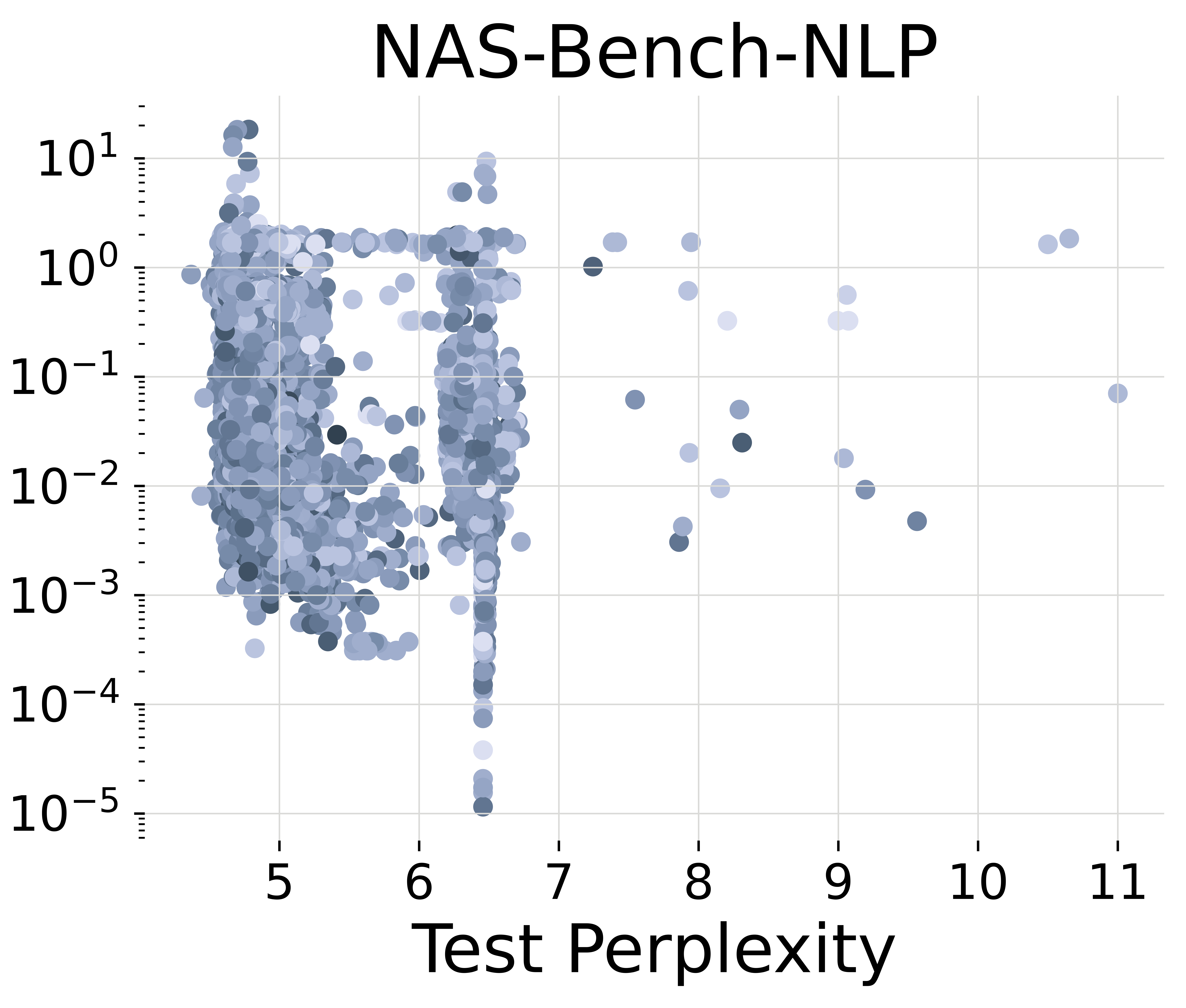}}
\caption{Zero-cost NAS \texttt{epsinas} metric performance illustration for NAS-Bench-101 search space, CIFAR-10 dataset  and NAS-Bench-NLP search space, PTB dataset. The horizontal axis shows test accuracy upon training. Each dot corresponds to an architecture; the darker the colour, the more parameters it contains. The figure shows $423{,}624$ and $14{,}322$ networks for NAS-Bench-101 and NAS-Bench-NLP, respectively (excluding architectures with \texttt{NaN} scores).}
\label{fig:results_101}
\end{center}
\end{figure*}

We use the NAS-Bench-101 search space to verify that the effectiveness of the \texttt{epsinas} metric, as demonstrated in the previous section, is not merely a result of overfitting to the NAS-Bench-201 space. It also allows us to assess its performance in a broader search space. As shown in Table \ref{tab:results_101} and Figure \ref{fig:results_101}, \texttt{epsinas} performs well on NAS-Bench-101, once again surpassing the \texttt{nparams} proxy.

% Table overall performance NAS-Bench-101
\begin{table*}[!t]
\caption{Zero-cost metrics performance evaluated on NAS-Bench-101 search space, CIFAR-10 dataset. Values from \cite{abdelfattah2021zero} are given for reference between brackets. We highlight the best-performing metrics in bold (p-values between \texttt{epsinas} and other metrics are below 0.001).}
\label{tab:results_101}
\begin{center}
\footnotesize
\renewcommand{\tabcolsep}{0.07cm}
\renewcommand{\arraystretch}{1.2}
\begin{tabular}{lrrrrp{0.1mm}rrrrrr}
\toprule
\noalign{\vskip 1 pt} 
\multirow{2}{*}{Metric} & \multicolumn{4}{c}{Spearman $\rho$} & & \multicolumn{2}{c}{Kendall $\tau$} & \multicolumn{2}{c}{Top-10\%/} & \multicolumn{2}{c}{Top-64/} \\
% \cline{2-5}\cline{7-8}
\noalign{\vskip 1.5pt} 
  & \multicolumn{2}{c}{global} & \multicolumn{2}{c}{top-10\%} & & \multicolumn{1}{c}{global} & \multicolumn{1}{c}{top-10\%} & \multicolumn{2}{c}{top-10\%}& \multicolumn{2}{c}{top-5\%} \\
\midrule
\noalign{\vskip 1pt}
\texttt{grad\_sign}    &  0.45  &  &  &  &  &  &  &  &  &  & \\
\texttt{synflow}    &  0.37  &        & \pmb{0.14} &          &  &  0.25  & \pmb{0.10} & 22.67 & (23) &  4 & (12) \\
\texttt{grad\_norm} & -0.20  &        & -0.05 &  (0.05)  &  & -0.14  & -0.03 &  1.98 &  (2) &  0 &  (0) \\
\texttt{grasp}      &  0.45  &        & -0.01 &          &  &  0.31  & -0.01 & 25.60 & (26) &  0 &  (6) \\
\texttt{snip}       & -0.16  &        &  0.01 &  (-0.01) &  & -0.11  &  0.00 &  3.43 &  (3) &  0 &  (0) \\
\texttt{fisher}     & -0.26  &        & -0.07 &  (0.07)  &  & -0.18  & -0.05 &  2.65 &  (3) &  0 &  (0) \\
\texttt{nparams}   & 0.43   &        &   &          &  & 0.31  &   &  &    &  &      \\  
\texttt{zico}   & \pmb{0.63}   &        &   &          &  & \pmb{0.46}  &   &  &    &  &      \\  
$\xi$\texttt{-gsnr}    &  0.615  &        &     &         &      &  0.434  &    &      &    &    &      \\
\texttt{epsinas}   & 0.62   &        &  0.12 &          &  & 0.44  &  0.08 & \pmb{40.33} &    & \pmb{10} &      \\  
\bottomrule
\end{tabular}
\end{center}
\end{table*}

% Figure, epsinas vs Accuracy, NAS-Bench-101

\subsection{NAS-Bench-NLP}
Both NAS-Bench-201 and NAS-Bench-101 are created to facilitate NAS in image recognition. They operate convolutional networks of very similar constitutions. To truly probe the generalisability of the \texttt{epsinas} metric, we test it on NAS-Bench-NLP. Both input data format and architecture type differ from the first two search spaces.

Unfortunately, \cite{abdelfattah2021zero} provides no data for NAS-Bench-NLP, disabling us from using their results for calculations. Therefore, in Table \ref{tab:results_nlp}, we give only values provided in the paper together with our \texttt{epsinas} metric (data for \texttt{ficher} is absent).  It’s important to note that, unlike accuracy, perplexity -- commonly used in language-related machine learning problems -- should be minimized. Thus, the signs of correlations with scoring metrics must be reversed, which differs from the results reported in \cite{abdelfattah2021zero}.

\begin{table*}[!t]
\caption{Zero-cost metrics performance evaluated on NAS-Bench-NLP search space, PTB dataset. We highlight the best-performing metrics in bold (p-values between \texttt{epsinas} and other metrics are below 0.001).}
\label{tab:results_nlp}
\begin{center}
\footnotesize
\renewcommand{\tabcolsep}{0.07cm}
\renewcommand{\arraystretch}{1.2}
\begin{tabular}{lrrp{0.1mm}rrrr}
\toprule
\noalign{\vskip 1 pt} 
\multirow{2}{*}{Metric} & \multicolumn{2}{c}{Spearman $\rho$} & & \multicolumn{2}{c}{Kendall $\tau$} & Top-10\%/ & Top-64/ \\
% \cline{2-3}\cline{5-6}
\noalign{\vskip 1pt} 
  & \multicolumn{1}{c}{global} &  \multicolumn{1}{c}{top-10\%}&  & global & top-10\% & top-10\% & top-5\% \\
\midrule
\noalign{\vskip 1pt} 
\texttt{synflow}    &     0.34    &     0.10    &  &  &  &     22    &  \\
\texttt{grad\_norm} &    -0.21    &     0.03    &  &  &  &     10    &  \\
\texttt{grasp}      &     0.16    &   0.55    &  &  &  &      4    &  \\
\texttt{snip}       &    -0.19    &    -0.02    &  &  &  &     10    &  \\
\texttt{epsinas}   &   \pmb{-0.34}    &    \pmb{-0.12}    &  &    -0.24   &    -0.08  &    \pmb{26.23}   &      13    \\
\bottomrule
\end{tabular}
\end{center}
\end{table*}

\subsection{Computational cost}
\begin{figure*}[!t]
\begin{center}
\includegraphics[width=5in]{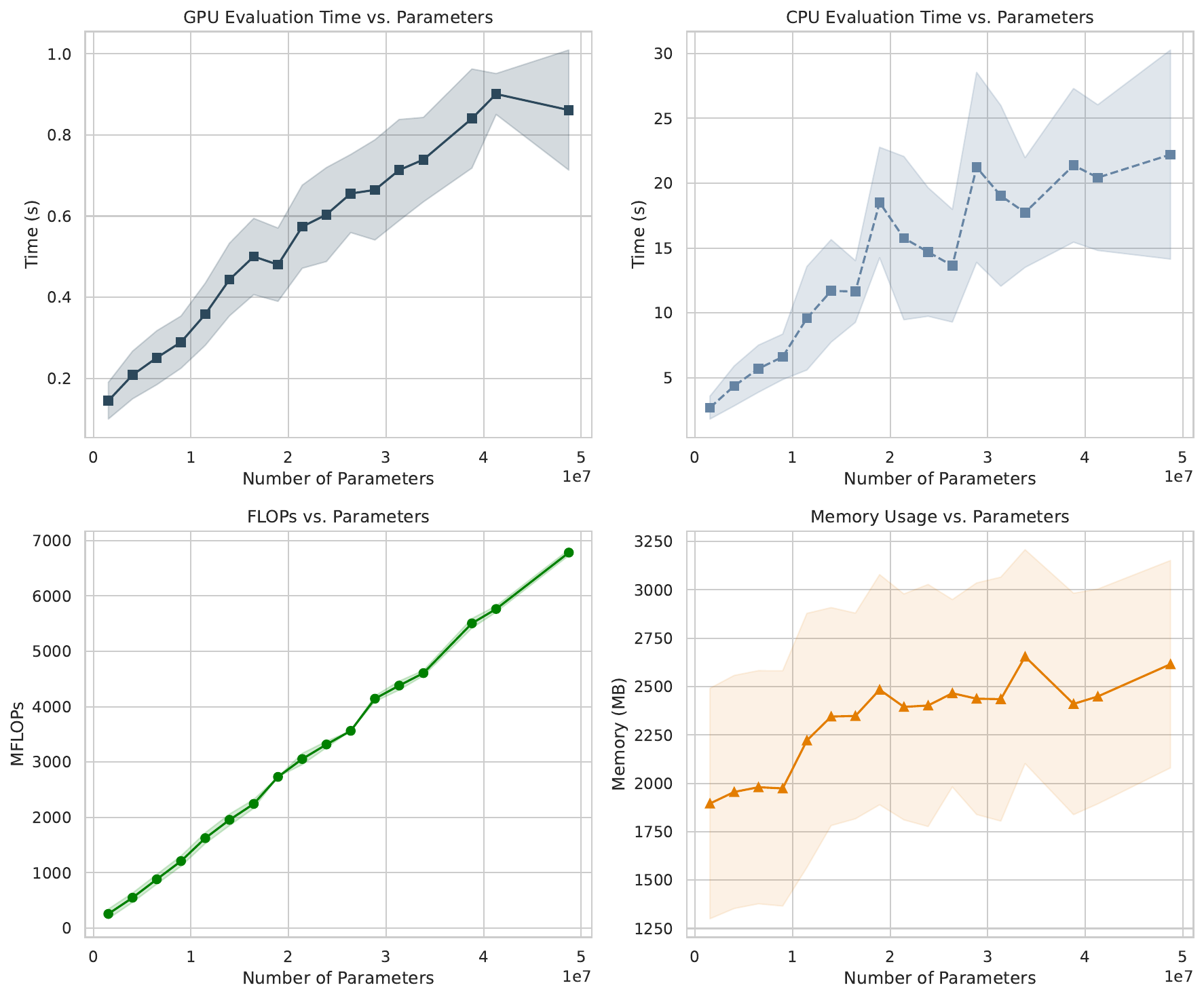}
\caption{Computational efficiency of the zero-cost NAS \texttt{epsinas} metric for the NAS-Bench-101 search space. The results are based on the evaluation of $10{,}000$ networks from the NAS-Bench-101 benchmark, divided into $20$ bins according to the number of parameters. The shaded area represents one standard deviation within each group. FLOPs are reported for a batch size of $1$, while memory usage and processing time are evaluated with a batch size of $256$.}
\label{fig:comp_perf}
\end{center}
\end{figure*}

One of the primary purposes of zero-shot proxies is to reduce the computational cost of the NAS process. Since \texttt{epsinas} only requires two forward passes through the data and does not involve gradient computation, it is efficient regarding both time and memory usage. Figure \ref{fig:comp_perf} presents the results of tests conducted using NAS-Bench-101. As anticipated, memory usage and processing time scale linearly with the number of parameters and FLOPs. It also demonstrates that \texttt{epsinas} can be computed using a CPU within 30 seconds for the largest NAS-Bench-101 networks consisting of several million parameters.

Comparing \texttt{epsinas} to other zero-cost metrics is not straightforward, as different methods use different search spaces, computational setups, and batch sizes. For instance, \texttt{zico} reports evaluation times using MobileNetV2 \cite{sandler2018mobilenetv2} with a budget of $1,000$ MFLOPs and a batch size of $128$. The authors noted that evaluating $10^5$ architectures required $10$ GPU hours.

MobileNetV2 networks range from $1.7$M to $6.9$M parameters, larger than NAS-Bench-201 but smaller than NAS-Bench-101. To provide a closer comparison, we evaluated \texttt{epsinas} on NAS-Bench-101 with a batch size of $128$, limiting architectures to $6.9$M parameters. Under these conditions, evaluating $10^5$ architectures took approximately $7$ GPU hours, with the \texttt{epsinas} computation requiring only $2$ GPU hours; the remaining time was spent on model loading.

While this is not a direct comparison, \texttt{epsinas} does not involve gradient computation, so it is reasonable to expect it to be faster per architecture evaluation. The results support this expectation.

\subsection{Correlation with \texttt{nparams}}
\begin{figure*}[!t]
\begin{center}
\includegraphics[width=3in]{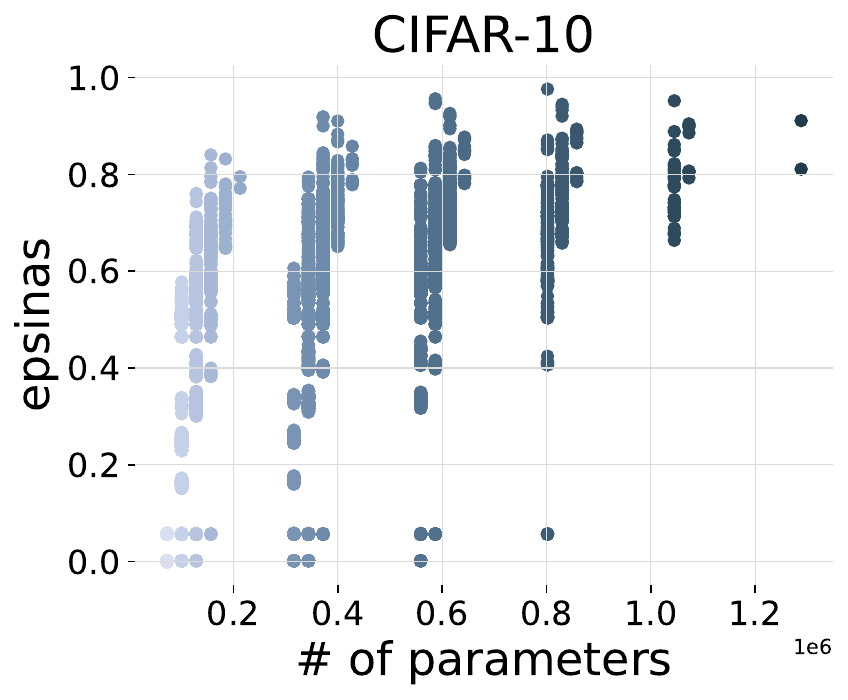}
\caption{Visualisation of the relationship between \texttt{epsinas} score and the number of parameters in a network, \texttt{nparams} for CIFAR-10 data of NAS-Bench-201 benchmark.}
\label{fig:nparams}
\end{center}
\end{figure*}

As noted in previous studies \citep{ning2021evaluating,li2023zico,white2022deeper}, one of the best naive evaluators of a network's performance is the number of parameters, \texttt{nparams}. Generally, the more parameters a network has, the better we expect its predictive performance to be. However, bigger does not always mean better: large networks are susceptible to overfitting and can be prohibitively expensive to train \citep{dauphin2013big, lecun1990optimal}. Moreover, research on network pruning has shown that, with the right architecture, smaller subnetworks can achieve comparable or even superior performance to their larger counterparts \citep{frankle2018lottery}. Hence, when seeking an optimal evaluation proxy, it is essential to ensure that it can effectively identify smaller networks. Figure \ref{fig:nparams} illustrates the relationship between \texttt{epsinas} and the number of parameters, showing no strong preference for larger network sizes.

\subsection{Integration with other NAS methods}
\label{sec:implement}
While zero-cost metrics are robust on their own, they benefit from being integrated with other NAS algorithms. \cite{lopes2022efficient} showed that combining a zero-cost metric based on Jacobian covariance, similar to \cite{mellor2020neural}, with REA \citep{real2019regularized} speeds up evolutionary search and helps find top-performing architectures. This work demonstrates how the \texttt{epsinas} metric can improve random search and ageing evolution algorithms.

This work follows the implementation of \cite{abdelfattah2021zero}.

\textbf{Random search}
Each new architecture is selected randomly and evaluated based on test accuracy. The final test performance is reported without using any auxiliary metrics.

\textbf{Random search with warm-up}
A warm-up pool of $3{,}000$ randomly sampled architectures is first ranked using \texttt{epsinas}. Next, models are "trained" in order of decreasing \texttt{epsinas} value by retrieving their test accuracies. Training stops once the budget of $300$ architectures is reached.

\textbf{Ageing evolution}
We implement the basic ageing evolution algorithm from \cite{real2019regularized}. The process starts with a randomly sampled pool of $64$ architectures. Parents are selected based on test accuracy, and a child is generated through mutation (with an edit distance of $1$). The child is added to the pool, while the oldest architecture is removed. This process repeats until $300$ architectures have been trained.

\textbf{Ageing evolution with warm-up}
A warm-up phase begins with $3{,}000$ sampled architectures, which are ranked using the \texttt{epsinas} metric. The $64$ architectures with the highest \texttt{epsinas} values form the initial pool. The standard ageing evolution process continues, where parents are selected based on test accuracy.

\textbf{Ageing evolution with move}
Instead of using test accuracy for parent selection, architectures are assessed based on their \texttt{epsinas} scores.

We conduct $100$ random runs for each algorithm and report the mean and standard deviation.

\begin{figure*}[!t]
    \centering
    \begin{minipage}{.32\linewidth}
        \subfigure{
        \includegraphics[width=.95\textwidth]{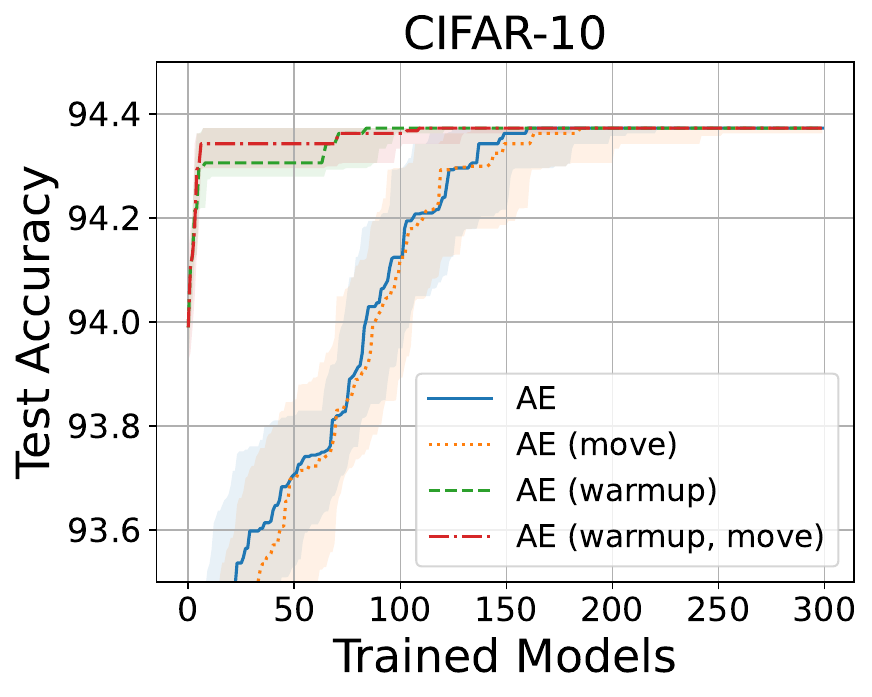}
        }\\
        \hfill
        \subfigure{
        \includegraphics[width=.95\textwidth]{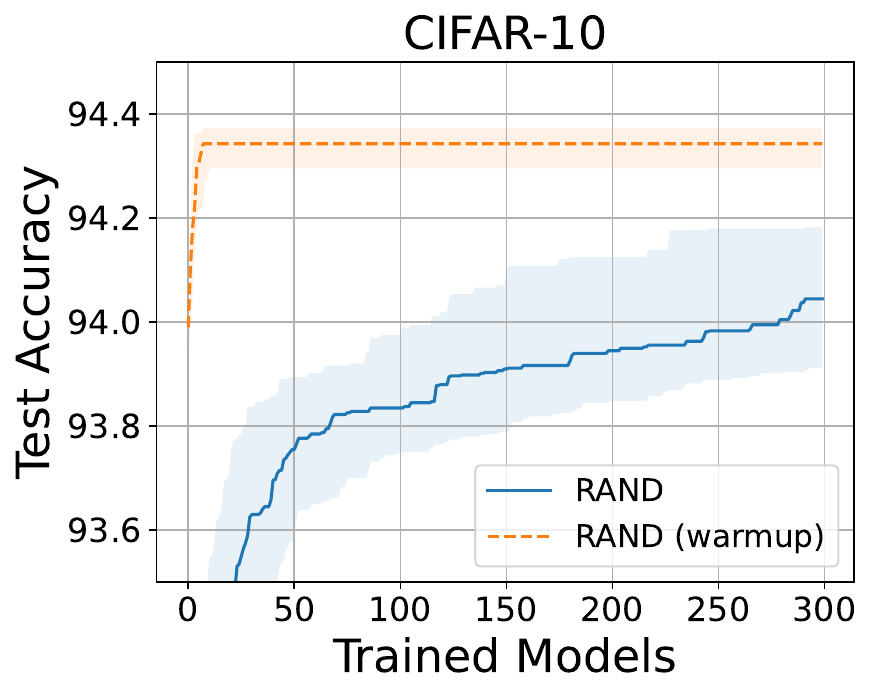}
        } 
    \end{minipage}
    \begin{minipage}{.3\linewidth}
        \subfigure{
        \includegraphics[width=.95\textwidth]{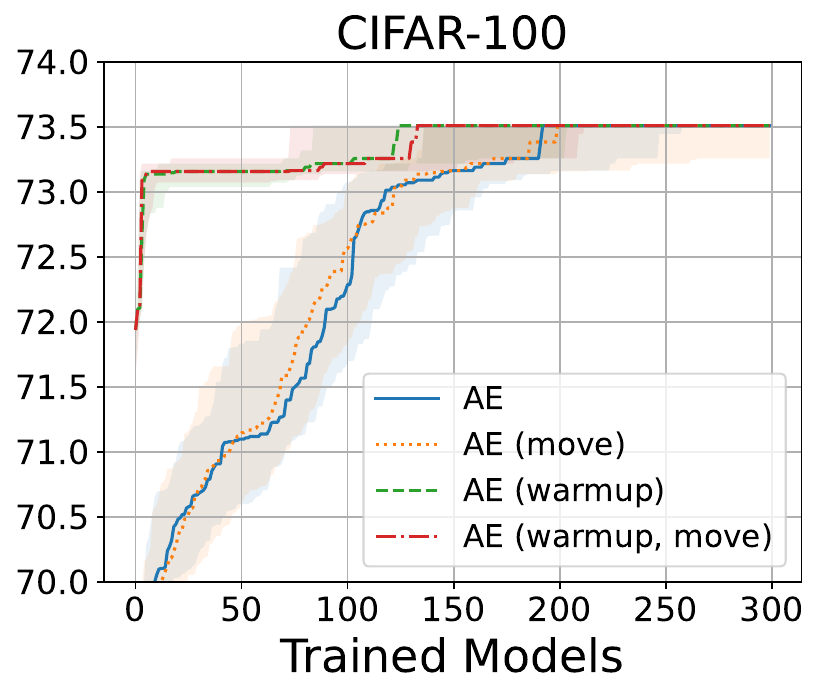}
        }\\
        \hfill
        \subfigure{
        \includegraphics[width=.95\textwidth]{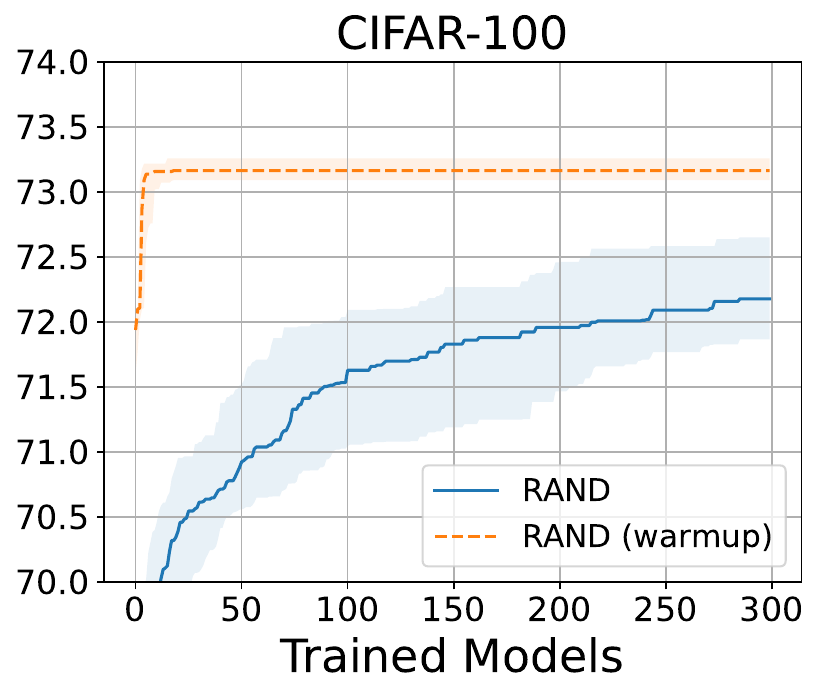}
        }
    \end{minipage}
    \begin{minipage}{.3\linewidth}
        \subfigure{
        \includegraphics[width=.95\textwidth]{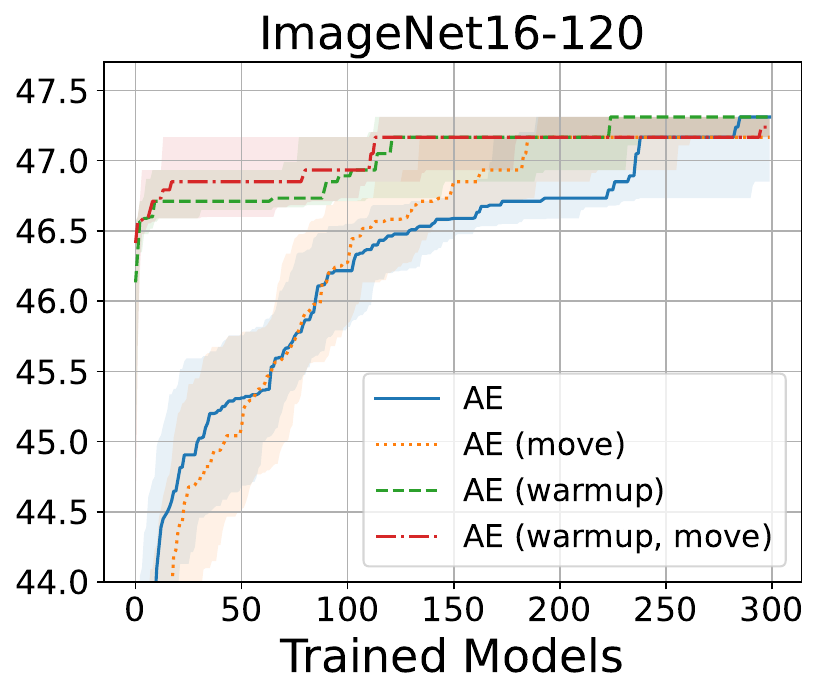}
        }\\
        \hfill
        \subfigure{
        \includegraphics[width=.95\textwidth]{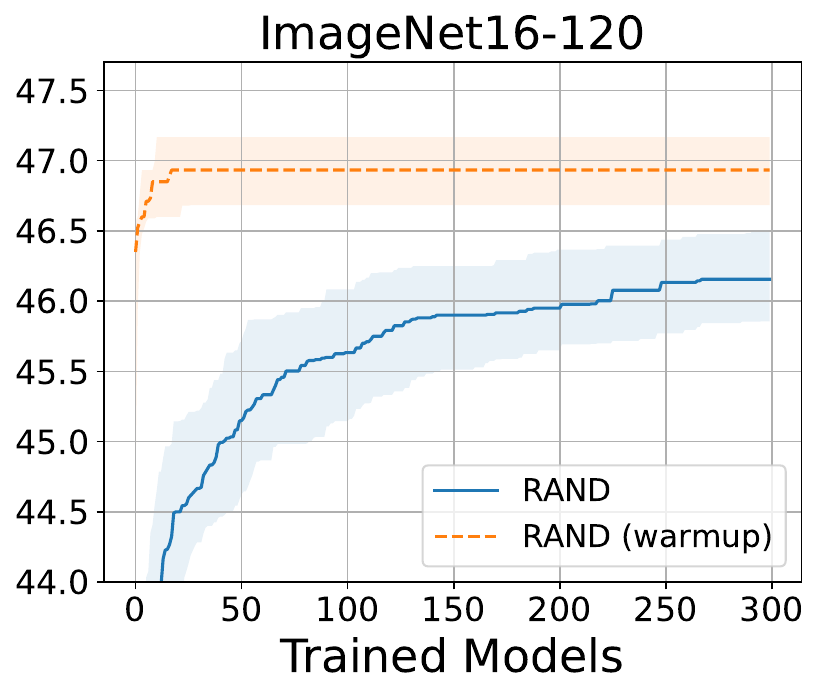}
        }
    \end{minipage}
\caption{\texttt{epsinas} integration within ageing evolution (top) and random search (bottom) NAS algorithms for three datasets from NAS-Bench-201 search space.} 
\label{fig:integration}
\end{figure*}

Figure \ref{fig:integration} demonstrates that the \texttt{epsinas} metric significantly improves both efficiency and accuracy, with the best results achieved when combined with a warm-up. Figure \ref{fig:integration_other} compares warm-up performances across various zero-cost metrics.

\section{Ablation studies}
\label{sec:ablations}
While our metric is straightforward to implement, it depends on several hyperparameters. In this section, we present and analyse the results of our ablation studies.

\subsection{Weights}

To compute the \texttt{epsinas} metric, we initialise the networks with two distinct constant shared weights. The question is: how do we fix their values, and how does this choice affect the whole method?

To answer this question, we ran a series of tests with multiple pairs of weights. It makes sense to set the first weight always greater than the second one in pairs. This consideration stems from the fact that \texttt{epsinas} is based on the mean absolute difference between two initialisations. Therefore, it will output the same score for [$w_{1}$, $w_{2}$] and [$w_{2}$, $w_{1}$] combinations and will be exactly zero in the case where the two weights are equal. Tables \ref{tab:weights_101}, \ref{tab:weights_201}, \ref{tab:weights_nlp} summarise the results of our tests. For the NAS-Bench-201 search space, the outcomes are very similar between the datasets, and the minor differences do not affect the conclusions drawn from this data.

% Tables, weights ablation
\begin{table*}[!b]
\vspace{-10pt}
\caption{Zero-cost metrics performance was evaluated using the NAS-Bench-101 search space and the CIFAR-10 dataset. We start with $5{,}000$ randomly picked architectures. The number of architectures with non-\texttt{NaN} values is given in the second column.}
\label{tab:weights_101}
\begin{center}
\footnotesize
\renewcommand{\tabcolsep}{0.07cm}
\renewcommand{\arraystretch}{1.2}
\begin{tabular}{llrrp{0.1mm}rrrc}
\toprule
\multirow{2}{*}{Weights} & \multirow{2}{*}{Archs} &\multicolumn{2}{c}{Spearman $\rho$} & & \multicolumn{2}{c}{Kendall $\tau$} & \multicolumn{1}{c}{Top-10\%/} & \multicolumn{1}{c}{Top-64/} \\
% \cline{3-4}\cline{6-7}
\noalign{\vskip 1.5pt} 
  & & \multicolumn{1}{c}{global} & \multicolumn{1}{c}{top-10\%} & & \multicolumn{1}{c}{global} & \multicolumn{1}{c}{top-10\%} & \multicolumn{1}{c}{top-10\%}& \multicolumn{1}{c}{top-5\%} \\
\midrule
0.0001, 0.001  & 2120  &  0.47  &  0.41 &  &  0.33  &  0.28  &  38.21 & 23 \\
0.0001, 0.01  &  2120  &  0.61  &  0.03 &  &  0.43  &  0.02  &  41.98 & 24 \\
0.0001, 0.1   &  2120  &  0.61  &  0.03 &  &  0.43  &  0.02  &  41.98 & 24 \\
0.0001, 1     &  2120  &  0.61  &  0.03 &  &  0.43  &  0.02  &  41.98 & 24 \\
0.001, 0.01   &  4968  &  0.29  & -0.05 &  &  0.20  & -0.03  &  17.30 & 10 \\
0.001, 0.1    &  4968  &  0.29  & -0.05 &  &  0.19  & -0.03  &  17.30 &  8 \\
0.001, 1      &  4968  &  0.29  & -0.05 &  &  0.19  & -0.03  &  17.30 &  8 \\
0.01, 0.1     &  5000  &  0.21  & -0.12 &  &  0.14  & -0.08  &  4.60  &  0 \\
0.01, 1       &  5000  &  0.21  & -0.12 &  &  0.14  & -0.08  &  4.60  &  0 \\
0.1, 1        &  5000  & -0.47  &  0.04 &  & -0.32  &  0.03  &  0.20  &  0 \\
\bottomrule
\end{tabular}
\end{center}
\end{table*}

\begin{table*}[!t]
\caption{Zero-cost metrics performance was evaluated in the NAS-Bench-201 search space and the CIFAR-10 dataset. We start with $5{,}000$ randomly picked architectures. The number of architectures with non-\texttt{NaN} values is given in the second column.}
\label{tab:weights_201}
\begin{center}
\footnotesize
\renewcommand{\tabcolsep}{0.07cm}
\renewcommand{\arraystretch}{1.2}
\begin{tabular}{llrrp{0.1mm}rrrc}
\toprule
\multirow{2}{*}{Weights} & \multirow{2}{*}{Archs} &\multicolumn{2}{c}{Spearman $\rho$} & & \multicolumn{2}{c}{Kendall $\tau$} & \multicolumn{1}{c}{Top-10\%/} & \multicolumn{1}{c}{Top-64/} \\
% \cline{3-4}\cline{6-7}
\noalign{\vskip 1.5pt} 
  & & \multicolumn{1}{c}{global} & \multicolumn{1}{c}{top-10\%} & & \multicolumn{1}{c}{global} & \multicolumn{1}{c}{top-10\%} & \multicolumn{1}{c}{top-10\%}& \multicolumn{1}{c}{top-5\%} \\
\midrule
1e-07, 1e-06  &  2957  &  0.33  &  -0.20  & & 0.22  & -0.13  &  4.76  &  1 \\
1e-07, 1e-05  &  2957  &  0.39  &  -0.06  & & 0.29  & -0.04  &  9.15  &  2 \\
1e-07, 0.0001 &  2957  &  0.47  &  -0.01  & & 0.32  &  0.00  &  9.12  &  2 \\
1e-07, 0.001  &  2957  &  0.62  &   0.06  & & 0.43  &  0.05  &  18.92 &  3 \\
1e-07, 0.01   &  2957  &  0.83  &   0.69  & & 0.64  &  0.49  &  68.40 & 56 \\
1e-07, 0.1    &  2957  &  0.84  &   0.64  & & 0.65  &  0.46  &  68.03 & 56 \\
1e-07, 1      &  2957  &  0.87  &   0.59  & & 0.70  &  0.43  &  65.88 & 45 \\
1e-06, 1e-05  &  2957  &  0.41  &  -0.06  & & 0.30  & -0.04  &  9.15  &  2 \\
1e-06, 0.0001 &  2957  &  0.47  &  -0.01  & & 0.32  &  0.00  &  9.12  &  2 \\
1e-06, 0.001  &  2957  &  0.62  &   0.06  & & 0.43  &  0.05  &  18.92 &  3 \\
1e-06, 0.01   &  2957  &  0.83  &   0.69  & & 0.64  &  0.49  &  68.40 & 56 \\
1e-06, 0.1    &  2957  &  0.84  &   0.64  & & 0.65  &  0.46  &  68.03 & 56 \\
1e-06, 1      &  2957  &  0.87  &   0.59  & & 0.70  &  0.43  &  65.88 & 45 \\
1e-05, 0.0001 &  2957  &  0.47  &  -0.01  & & 0.32  &  0.00  &  9.12  &  2 \\
1e-05, 0.001  &  2957  &  0.62  &   0.06  & & 0.43  &  0.05  &  18.92 &  3 \\
1e-05, 0.01   &  2957  &  0.83  &   0.69  & & 0.64  &  0.49  &  68.40 & 56 \\
1e-05, 0.1    &  2957  &  0.84  &   0.64  & & 0.65  &  0.46  &  68.03 & 56 \\
1e-05, 1      &  2957  &  0.87  &   0.59  & & 0.70  &  0.43  &  65.88 & 45 \\
0.0001, 0.001 &  2957  &  0.62  &   0.06  & & 0.43  &  0.05  &  19.26 &  3 \\
0.0001, 0.01  &  2957  &  0.82  &   0.69  & & 0.63  &  0.49  &  68.42 & 56 \\
0.0001, 0.1   &  2957  &  0.84  &   0.64  & & 0.65  &  0.46  &  68.14 & 56 \\
0.0001, 1     &  2957  &  0.87  &   0.58  & & 0.69  &  0.42  &  65.88 & 46 \\
0.001, 0.01   &  4802  &  0.25  &   0.41  & & 0.17  &  0.28  &  30.35 & 52 \\
0.001, 0.1    &  4802  &  0.31  &   0.27  & & 0.20  &  0.19  &  26.82 & 25 \\
0.001, 1      &  4802  &  0.31  &   0.22  & & 0.21  &  0.15  &  26.26 &  6 \\
0.01, 0.1     &  4907  &  0.15  &   0.42  & & 0.10  &  0.29  &  37.47 & 57 \\
0.01, 1       &  4907  &  0.12  &   0.39  & & 0.06  &  0.27  &  35.85 & 57 \\
0.1, 1        &  4907  &  0.01  &   0.26  & &-0.03  &  0.16  &  0.84  &  0 \\
\bottomrule
\end{tabular}
\end{center}
\end{table*}

\begin{table*}[!t]
\caption{Zero-cost metrics performance was evaluated using the NAS-Bench-NLP search space and the PTB dataset. We start with $5{,}000$ randomly picked architectures. The number of architectures with non-\texttt{NaN} values is given in the second column.}
\label{tab:weights_nlp}
\begin{center}
\footnotesize
\renewcommand{\tabcolsep}{0.07cm}
\renewcommand{\arraystretch}{1.2}
\begin{tabular}{llrrp{0.1mm}rrrc}
\toprule
\multirow{2}{*}{Weights} & \multirow{2}{*}{Archs} &\multicolumn{2}{c}{Spearman $\rho$} & & \multicolumn{2}{c}{Kendall $\tau$} & \multicolumn{1}{c}{Top-10\%/} & \multicolumn{1}{c}{Top-64/} \\
% \cline{3-4}\cline{6-7}
\noalign{\vskip 1.5pt} 
  & & \multicolumn{1}{c}{global} & \multicolumn{1}{c}{top-10\%} & & \multicolumn{1}{c}{global} & \multicolumn{1}{c}{top-10\%} & \multicolumn{1}{c}{top-10\%}& \multicolumn{1}{c}{top-5\%} \\
\midrule
1e-07,  1e-06  & 2393  & -0.24 &  0.32  & & -0.16 &  0.22  &  0.00  & 0 \\
1e-07,  1e-05  & 2391  & -0.22 &  0.52  & & -0.15 &  0.39  &  1.67  & 0 \\
1e-07,  0.0001 & 2388  & -0.35 &  0.49  & & -0.24 &  0.37  &  0.42  & 1 \\
1e-07,  0.001  & 2332  & -0.42 &  0.50  & & -0.29 &  0.37  &  2.99  & 1 \\
1e-07,  0.01   & 1215  & -0.30 &  0.13  & & -0.21 &  0.09  &  4.20  & 1 \\
1e-07,  0.1    &  585  & -0.49 &  0.55  & & -0.36 &  0.44  &  1.69  & 1 \\
1e-07,  1      &  403  & -0.12 &  0.08  & & -0.08 & -0.00  & 12.20  & 7 \\
1e-06,  1e-05  & 3286  & -0.11 &  0.49  & & -0.07 &  0.36  &  0.91  & 0 \\
1e-06,  0.0001 & 3283  & -0.24 &  0.47  & & -0.16 &  0.34  &  0.62  & 0 \\
1e-06,  0.001  & 3226  & -0.33 &  0.44  & & -0.23 &  0.33  &  4.64  & 1 \\
1e-06,  0.01   & 1639  & -0.26 &  0.13  & & -0.18 &  0.09  &  3.40  & 2 \\
1e-06,  0.1    &  841  & -0.41 &  0.13  & & -0.28 &  0.12  & 10.71  & 1 \\
1e-06,  1      &  594  & -0.09 &  0.13  & & -0.06 &  0.08  & 11.67  & 7 \\
1e-05,  0.0001 & 3948  & -0.32 &  0.30  & & -0.21 &  0.22  &  0.51  & 0 \\
1e-05,  0.001  & 3892  & -0.35 &  0.20  & & -0.24 &  0.16  &  3.85  & 2 \\
1e-05,  0.01   & 1900  & -0.32 &  0.27  & & -0.22 &  0.18  &  2.86  & 1 \\
1e-05,  0.1    &  913  & -0.38 &  0.31  & & -0.28 &  0.25  & 13.04  & 1 \\
1e-05,  1      &  649  & -0.18 &  0.36  & & -0.12 &  0.25  & 10.77  & 7 \\
0.0001,  0.001 & 3987  & -0.35 &  0.18  & & -0.24 &  0.14  &  3.76  & 0 \\
0.0001,  0.01  & 1943  & -0.37 &  0.38  & & -0.26 &  0.25  &  3.59  & 1 \\
0.0001,  0.1   &  925  & -0.43 &  0.26  & & -0.30 &  0.21  & 11.83  & 1 \\
0.0001,  1     &  663  & -0.26 &  0.35  & & -0.18 &  0.24  & 10.45  & 7 \\
0.001,  0.01   & 1936  & -0.32 &  0.40  & & -0.21 &  0.26  &  2.59  & 3 \\
0.001,  0.1    &  918  & -0.38 &  0.27  & & -0.30 &  0.22  &  8.70  & 2 \\
0.001,  1      &  663  & -0.23 & -0.20  & & -0.19 & -0.17  &  7.46  & 5 \\
0.01,  0.1     &  910  & -0.36 &  0.04  & & -0.25 &  0.02  &  8.79  & 2 \\
0.01,  1       &  654  & -0.25 &  0.06  & & -0.16 &  0.04  &  9.09  & 5 \\
0.1,  1        &  652  & -0.27 &  0.20  & & -0.19 &  0.13  & 10.61  & 7 \\
\bottomrule
\end{tabular}
\end{center}
\end{table*}

The results indicate that variations in initialisation do affect performance. Two key factors influence this impact. First, a more significant difference between the two weights leads to a higher correlation between \texttt{epsinas} and trained performance. It is intuitive -- when the weights are too similar, the output difference becomes minimal, making it difficult to distinguish between architectures based on \texttt{epsinas}.

Secondly, we can see that too low and too high weights result in fewer architectures with non-\texttt{NaN} scores. Extreme weights result in all zeros or infinite output values and, consequently, in \texttt{NaN} value of the \texttt{epsinas} metric. Naturally, the deeper the network is, the more signal attenuation is, and the higher the probability of \texttt{NaN} score. We remember that before \texttt{epsinas} computation, the outputs are normalised to the same range $[0, 1]$, which means the metric value per se does not become smaller for deeper networks. This extinction effect is due to the limited sensitivity of \texttt{float} type. Theoretically, with infinite float sensitivity, there would be no extinction and hence no effects on the choice of the weights.

% Table, optimal weights
\begin{table*}
\caption{Optimal weights for epsinas evaluation for three search spaces and their datasets.}
\label{tab:optimal_weights}
\begin{center}
\footnotesize
\renewcommand{\tabcolsep}{0.07cm}
\renewcommand{\arraystretch}{1.2}
\begin{tabular}{lll}
\toprule
Search space & Dataset & Optimal weights \\
\midrule
NAS-Bench-101 & CIFAR-10 & $[10^{-4} , 10]$ \\
\multirow{3}{*}{NAS-Bench-201} & CIFAR-10 & $[10^{-7} , 1]$ \\
 & CIFAR-100 & $[10^{-7} , 1]$ \\
  & ImageNet16-120 & $[10^{-7} , 1]$ \\
\noalign{\vskip 1 pt} 
NAS-Bench-NLP & PTB & $[10^{-5} , 10^{-3}]$ \\
\bottomrule
\end{tabular}
\end{center}
\end{table*}

As a rule of thumb, for every new ML problem, we suggest running \texttt{epsinas} evaluation on a subset of architectures with several weights and selecting the minimum and maximum weights that do not cause excessive \texttt{NaN} outputs. While we acknowledge that this procedure is subjective, importantly, it does not require any training.

Weight combinations that we selected for each search space are given in Table \ref{tab:optimal_weights}.

\subsection{Initialisation algorithm}
The choice to initialise the weights with constant values provides a ground for discussion. On one hand, using constant shared weights ensures a consistent initialisation across different structures, eliminating the noise introduced by random initialisations. It allows us to isolate and analyse the impact of neural geometry more effectively. Averaging results over multiple runs with different random seeds, as in \cite{gracheva2021trainless}, can reduce the randomness noise but increases computational time proportionally. 

Table \ref{tab:init_ablation} presents the results of our ablation studies, where \texttt{epsinas} is computed using different initialisation schemes. The correlation coefficients are based on the \texttt{epsinas} scores from two initialisations but using standard initialisation methods instead of constant shared weights. For every initialisation scheme, we only modify the convolutional layers. The rest of the architecture is initialised as per the original benchmark space: constant initialisation with ones for the batch normalisation, normal uniform for the linear layer, and biases set to zero.

% Table, initialisation ablation
\begin{table*}[!t]
\caption{Initialisations ablation studies. The metric is computed over the first $5{,}000$ architectures in the NAS-Bench-201 benchmark.}
\label{tab:init_ablation}
\begin{center}
\footnotesize
\renewcommand{\tabcolsep}{0.07cm}
\renewcommand{\arraystretch}{1.2}
\begin{tabular}{lrrp{0.1mm}rrrc}
\toprule
\multirow{2}{*}{Initialisation} &  \multicolumn{2}{c}{Spearman $\rho$} & & \multicolumn{2}{c}{Kendall $\tau$} & Top-10\%/ & Top-64/ \\
% \cline{3-4}\cline{6-7}
\noalign{\vskip 1.5pt} 
  & global & top-10\% & & global & top-10\% & top-10\%& top-5\% \\
\midrule
Uniform          & -0.02 & -0.01  &  &  -0.01 & -0.00  &  7.13   &  13  \\
Normal           & 0.08  & -0.20  &  &  0.05  & -0.13  &  15.83  &  4  \\
Kaiming uniform  & 0.07  & -0.17  &  &  0.05  & 0.12   &  13.60  &  2 \\
Kaiming normal   & 0.07  & -0.13  &  &  0.05  & -0.09  &  12.49  &  1  \\
Orthogonal       & 0.07  & -0.16  &  &  0.05  & -0.10  &  11.71  &  6  \\
\bottomrule
\end{tabular}
\end{center}
\end{table*}

\subsection{Batch size}
To test the method's sensitivity against the batch size, we ran the scoring routine with $8$ different batch sizes $[8, 16, 32, 64, 128, 512, 1024]$, each with $10$ random batches. The tests are done on the NAS-Bench-201 search space with its three datasets. We report medians with $25$ and $75$ percentiles over $10$ runs for each batch. We evaluate \texttt{epsinas} on $5{,}000$ architectures.

\begin{figure*}[!t]
    \centering
    \begin{minipage}{.32\linewidth}
        \subfigure{
        \includegraphics[width=.95\textwidth]{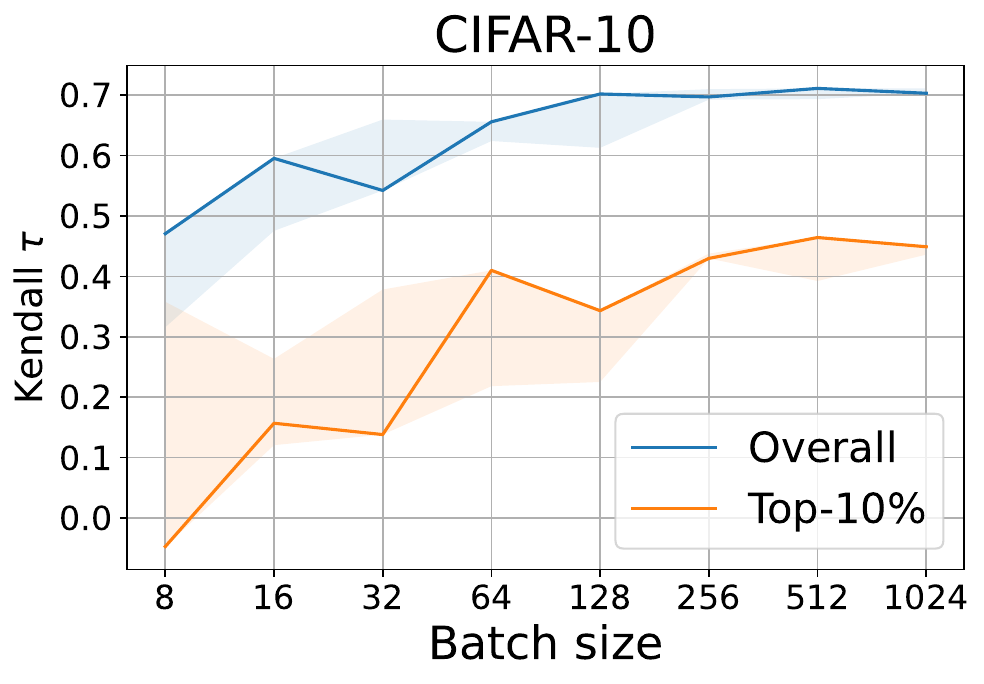}
        }\\
        \hfill
        \subfigure{
        \includegraphics[width=.95\textwidth]{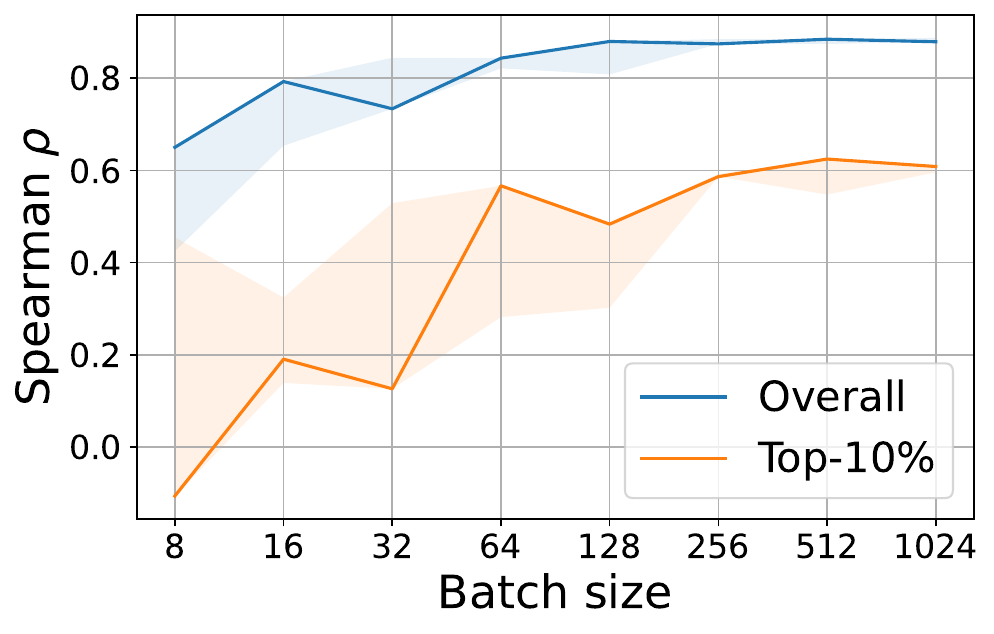}
        } 
    \end{minipage}
    \begin{minipage}{.32\linewidth}
        \subfigure{
        \includegraphics[width=.95\textwidth]{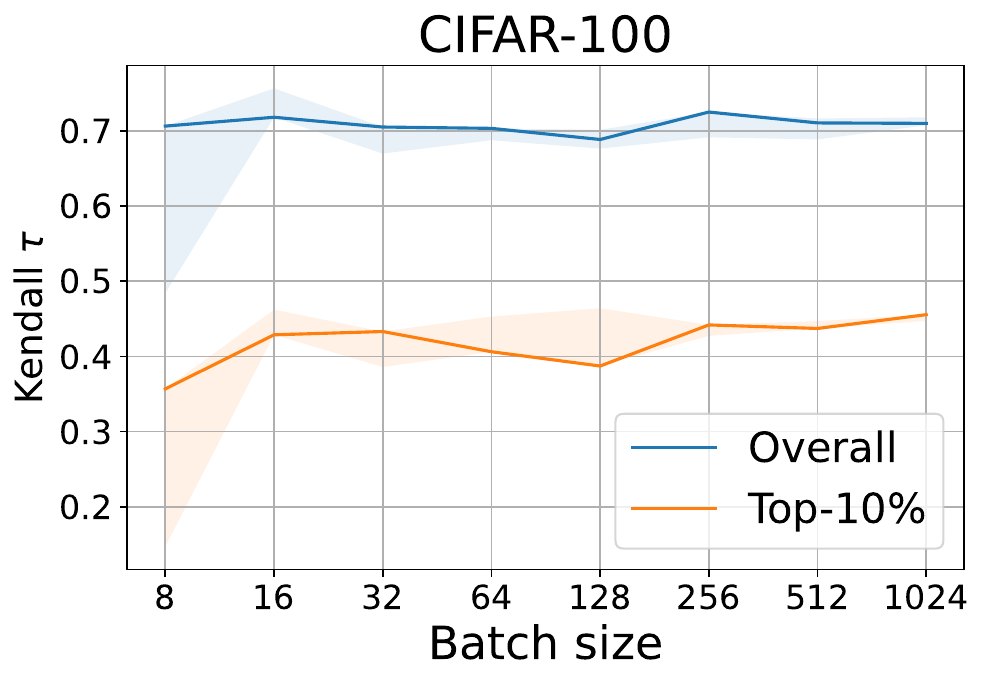}
        }\\
        \hfill
        \subfigure{
        \includegraphics[width=.95\textwidth]{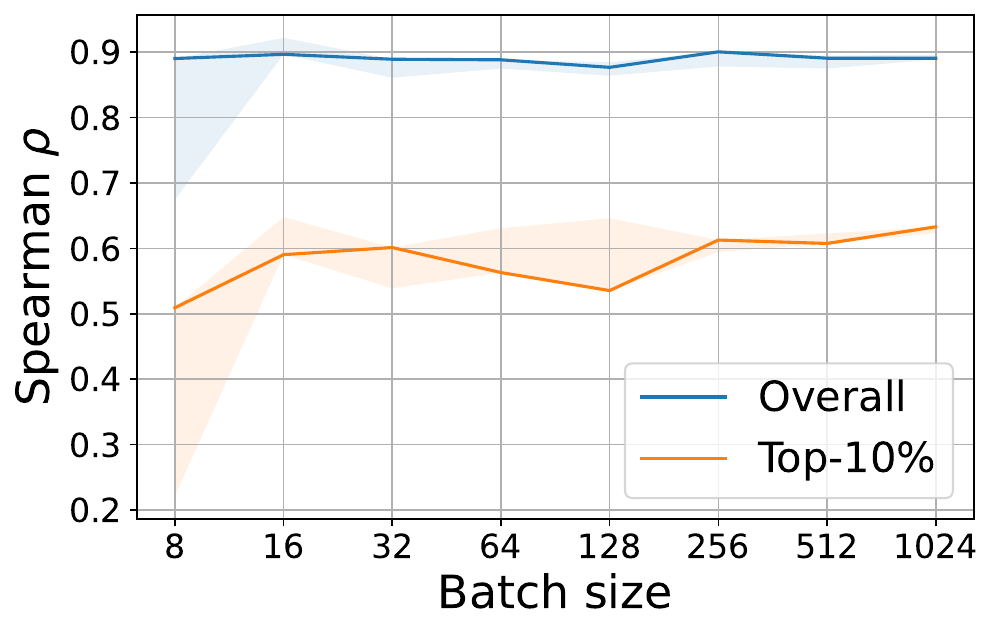}
        }
    \end{minipage}
    \begin{minipage}{.32\linewidth}
        \subfigure{
        \includegraphics[width=.95\textwidth]{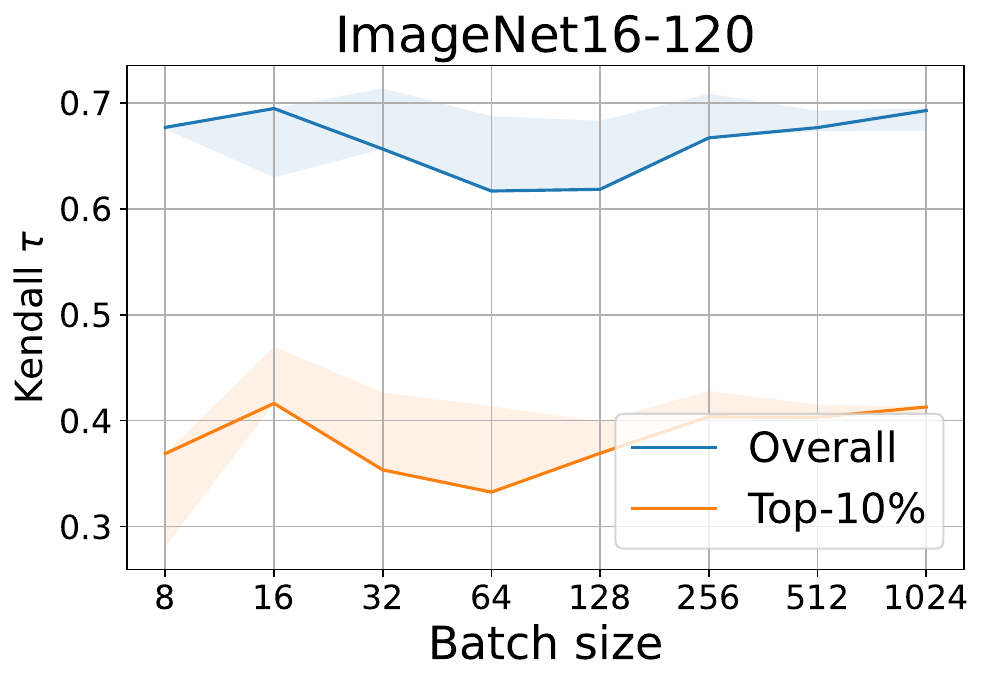}
        }\\
        \hfill
        \subfigure{
        \includegraphics[width=.95\textwidth]{figures/ablations/batch_size/BatchSize_Kendall_201_CIFAR10.pdf}
        }
    \end{minipage}
\caption{Batch size ablation study for NAS-Bench-201 search space, CIFAR-10 (left), CIFAR-100 (centre) and ImageNet16-120 (right) datasets.} 
\label{fig:ablation_batch}
\end{figure*}

Figure \ref{fig:ablation_batch} shows that, as expected, the larger the batch size, the better and more stable the performance of \texttt{epsinas} metric. There is almost no improvement for batches over $256$. We set the batch size to this value in the experiments throughout the paper.

\subsection{Embedding initialisation}
\label{sec:embedding}
To verify the effect of the embedding initialisation, we have run tests with $6$ different initialisations: 
\begin{itemize}
\item uniform positive 0.1: random uniform from ranges [0, 0.1]
\item uniform positive 1: random uniform from ranges [0, 1]
\item uniform centred 0.1: random uniform from ranges [-0.1, 0.1]
\item uniform centred 1: random uniform from ranges [-1, 1]
\item random 0.1: random normal centred at $0$ with a standard deviation of 0.1
\item random 1: random normal centred at $0$ with a standard deviation of 1        
\end{itemize}

Table \ref{tab:embed_ablation} summarises our results of the embedding ablations. Ablation is done on the NAS-Bench-NLP search space, PTB dataset (the only search space implementing embedding).

% Table, embedding ablation
\begin{table*}[!t]
\caption{Embedding ablation studies. Metric computed over $5{,}000$ initial architectures. The number of remaining architectures is given in the second column.}
\label{tab:embed_ablation}
\begin{center}
\footnotesize
\renewcommand{\tabcolsep}{0.07cm}
\renewcommand{\arraystretch}{1.2}
\begin{tabular}{llrrp{0.1mm}rrrc}
\toprule
\multirow{2}{*}{Embedding} & \multirow{2}{*}{Archs} &\multicolumn{2}{c}{Spearman $\rho$} & & \multicolumn{2}{c}{Kendall $\tau$} & \multicolumn{1}{c}{Top-10\%/} & \multicolumn{1}{c}{Top-64/} \\
% \cline{3-4}\cline{6-7}
\noalign{\vskip 1.5pt} 
  & & \multicolumn{1}{c}{global} & \multicolumn{1}{c}{top-10\%} & & \multicolumn{1}{c}{global} & \multicolumn{1}{c}{top-10\%} & \multicolumn{1}{c}{top-10\%}& \multicolumn{1}{c}{top-5\%} \\
\midrule
Uniform positive 0.1  &  782 & -0.38 & 0.17  &  &  -0.26 & 0.15  &  3.80  &  2 \\
Uniform positive 1    &  776 & -0.37 & 0.21  &  &  -0.26 & 0.18  &  2.56  &  2 \\
Uniform centered 0.1  &  783 & -0.40 & 0.18  &  &  -0.28 & 0.15  &  2.53  &  2 \\
Uniform centered 1    &  782 & -0.46 & 0.18  &  &  -0.33 & 0.16  &  1.27  &  1 \\
Random 0.1            &  783 & -0.42 & 0.18  &  &  -0.29 & 0.15  &  2.53  &  2 \\
Random 1              &  782 & -0.47 & 0.19  &  &  -0.33 & 0.16  &  1.27  &  1 \\
\bottomrule
\end{tabular}
\end{center}
\end{table*}

Care should be taken when initialising the networks containing the embedding layer: if embedding is initialised with all constants, there is no difference between the embedded input. In this case, our metric's performance will be analogous to that with a batch size of one.

Our ablation studies show that it does not significantly influence the outcomes as soon as the embedding is initialised with non-constant weights.

\subsection{Synthetic data}
\label{sec:synthetic_data}
In this section, we test the importance of the input data by feeding our metric several synthetic input data on CIFAR-10. We feed networks with $5$ types of data:
\begin{itemize}
\itemsep-0.2em 
\item actual data: a batch of CIFAR-10 data
\item grey scale images: images within the batch are solid colour ranging from black to white
\item random normal: images are filled with random values following a normal distribution with $[\mu, \sigma]=[0,1]$
\item random uniform: images are filled with random values following uniform distribution with $[\mu, \sigma]=[-1,1]$ 
\item random uniform (+): images are filled with random values following uniform distribution with $[min, \max]=[0,1]$
\end{itemize}

All the tests are performed with batch size of $256$, weights $[10^{-7}, 1]$ and $5{,}000$ architectures. Table \ref{tab:synthetic_data} shows that even though the performance with synthetic data drops compared to actual data, it is still reasonably good. 

Curiously, greyscale images, filled with constant values, show the closest results to the actual data. Note that \texttt{epsinas} metric with greyscale data outperforms \texttt{synflow}. It is an essential achievement because \texttt{synflow} does not use input data and is, therefore, data independent. Our results show that \texttt{epsinas} has the potential to be used with no data whatsoever.

% Table, synthetic data ablation
\begin{table*}[!t]
\caption{Synthetic data tests with CIFAR-10 dataset, NAS-Bench-201 search space. Tests are based on the evaluation of $5{,}000$ architectures.}
\label{tab:synthetic_data}
\begin{center}
\footnotesize
\renewcommand{\tabcolsep}{0.07cm}
\renewcommand{\arraystretch}{1.2}
\begin{tabular}{lrrp{0.1mm}rrrc}
\toprule
\multirow{2}{*}{Metric} & \multicolumn{2}{c}{Spearman $\rho$} & & \multicolumn{2}{c}{Kendall $\tau$} & Top-10\%/ & Top-64/ \\
% \cline{2-3}\cline{5-6}
\noalign{\vskip 1.5pt} 
  & \multicolumn{1}{c}{global} &  \multicolumn{1}{c}{top-10\%}&  & \multicolumn{1}{c}{global} & \multicolumn{1}{c}{top-10\%} & top-10\% & top-5\% \\
\midrule
\noalign{\vskip 1pt} 
Real data          & 0.87 & 0.59  & & 0.70 & 0.43 & 65.88 & $45$ \\
Grey scale          & 0.87 & 0.44 & & 0.68 & 0.31 & 59.46 & $34$ \\
Random normal       & 0.54 & 0.15 & & 0.38 & 0.14 & 14.97 & $3$  \\
Random uniform      & 0.56 & 0.17 & & 0.40 & 0.16 & 14.19 & $3$  \\
Random uniform (+) & 0.61 & 0.17  & & 0.43 & 0.16 & 16.22 & $4$  \\
\bottomrule
\end{tabular}
\end{center}
\end{table*}

\section{Discussion}
Our experiments show that despite its simplicity, \texttt{epsinas} metric shows surprisingly good performance across different benchmark datasets. It shows the most consistent performance with NAS-Bench-201, suggesting that this benchmark defines a relatively easy-to-navigate search space. Tables \ref{tab:results_201}, \ref{tab:results_101}, \ref{tab:results_nlp} demonstrates that \texttt{epsinas} outperforms most of other zero-cost metrics in terms of Spearman and Kendal correlations. The differences are statistically significant, with p-values between \texttt{epsinas} and other metrics below 0.001.

Notably, we highlight the key differences between our previous work (\texttt{cv} metric, \cite{gracheva2021trainless}) and the current approach. The two methods differ in how they de-emphasize weights and the type of outputs used for statistical computation. Using two fixed weight initialisations more effectively isolates the influence of network geometry compared to averaging over a hundred random initialisations. Additionally, the \texttt{epsinas} metric is significantly more efficient, reducing computation time by a factor of fifty.

We also compare the performance of \texttt{epsinas} with its conceptual counterparts, \texttt{zico} and $\xi$\texttt{-gsnr}. While the metrics are based on similar principles, \texttt{zico} and $\xi$\texttt{-gsnr} use gradients instead of raw outputs, which appears to limit their performance. This difference may also be attributed to random initialisation, as opposed to the shared constant weight initialisation used by \texttt{epsinas}. In all cases, \texttt{epsinas} consistently outperforms the naive \texttt{nparams} proxy, which relies solely on the total number of parameters in a given network.

It is important to note that we used NAS-Bench-201 to develop our metric, so the findings from this part of the study may be biased. To test the generalizability of the metric, we applied it to the NAS-Bench-101 benchmark. As shown in Table \ref{tab:results_101}, \texttt{epsinas} continues to demonstrate strong performance. The peculiar vertical lines in the results are due to clustering based on test accuracy. Since each architecture in NAS-Bench-101 is trained with three random seeds, the four clusters represent cases where training diverges for certain random initialisations. Using more random seeds would likely reveal a smoother pattern and stronger correlations.

By design, we expect our metric to be independent of the specific neural architecture structure. To confirm this claim, we show its applicability to architectures to RNN models with NAS-Bench-NLP.

While the statistics in Table \ref{tab:results_nlp} may not seem extraordinary at first glance, Figure \ref{fig:results_101} demonstrates a visible trend of improved architecture selection as the \texttt{epsinas} metric value increases. This positive trend is particularly noteworthy, especially considering the noise inherent to the benchmark. Factors such as the limited sample size of networks within a vast architecture space, variations in training hyperparameters, dropout rates, and other variables can contribute to performance inconsistencies in NAS. Despite these challenges, the trend is strong and consistent enough to confidently conclude that the \texttt{epsinas} metric shows great potential for effectively evaluating recurrent-type architectures.

While \texttt{epsinas} metric shows solid empirical performance, the underlying reasons for this remain unclear.

There are several hints towards its understanding. First, mathematically, \texttt{epsinas} represents the difference in the output distribution shapes between initialisations. The output shape is affected by layer widths, activation functions, batch normalisation, skip connections and other factors, which we generally call network geometry. By setting weights to a constant shared value, one can probe the effects of the geometry without being obstructed by the randomness of initialisation.

Second, according to the weight ablation studies (Section \ref{sec:ablations}), the best performance is achieved when we set the weights to the lowest and highest values that do not lead to excessive output explosion or vanishing. Therefore, \texttt{epsinas} measures the amplitude of the outputs' distribution shape change due to geometry.

Finally, we see that grey-scale solid images work reasonably well as inputs during the synthetic data studies. The distribution over the input samples is uniform, which makes it easier to track the changes in the distribution shape as the signal propagates through the network.

\subsection{Future directions}
While the presented \texttt{epsinas} metric demonstrates solid performance, several open questions remain. One of the key directions for future development is to assess its applicability to more types of neural topologies, such as transformer \cite{chen2021glit, zhou2024training}, generative adversarial \cite{gao2020cvpr} or graph neural networks \cite{ru2020neural}, and to test it with hierarchical search spaces \citep{christoforidis2021novel, ru2020neural, liu2019cvpr, chen2021glit}. It is also interesting to evaluate the metric in combination with other advanced NAS algorithms, such as multi-objective optimization \citep{zhao2021multi}, differentiable architecture search \cite{xiang2023zero}, Bayesian optimization \citep{snoek2012practical, shen2023proxybo}, or within the Parametric Zero-Cost Proxies \citep{dong2024parzc} framework.

The metric's performance is influenced by various hyperparameters, including weight values, batch size, and initialization scheme. While we conducted ablation studies on their individual effects, potential interactions between these hyperparameters remain unexplored, making it important to investigate their joint impact. Additionally, the current weight setup is based on empirical choices and would benefit from a solid theoretical foundation.

A deeper theoretical study of the metric is particularly needed given its similarities to the \texttt{zico} metric. For instance, it remains unclear why the mean absolute error over raw outputs outperforms the deviation over gradients and how this behaviour relates to the initialization scheme. Furthermore, examining the performance of the \texttt{zico} metric when initialized with constant shared weights could provide further insights.

\section{Conclusion}
This work presents a simple zero-cost NAS scoring metric, \texttt{epsinas}, which measures how much a network's outputs change when initialized with two different constant shared weights. We compute it by performing two forward passes and calculating the normalized MAE between the raw outputs.

For a simple metric, \texttt{epsinas} shows an unexpectedly high correlation with network performance in CNNs and RNNs, outperforming most existing zero-shot NAS proxies. We evaluate the \texttt{epsinas} metric on three widely-used NAS search spaces: NAS-Bench-201, NAS-Bench-101, and NAS-Bench-NLP. Our results show strong performance across these search spaces. 

The metric is computationally efficient, requiring only a fraction of a GPU second to evaluate a single architecture, with minimal memory demand, as it does not require gradient computation or labelled data. \texttt{epsinas} can be run on a CPU, making it broadly accessible. While small to medium-sized networks can be processed efficiently on a CPU, larger architectures may require GPU acceleration to handle forward passes more quickly, especially when evaluating multiple candidate architectures in succession. Optimizing CPU-GPU workload distribution could help balance energy consumption and speed in resource-constrained environments such as mobile and embedded systems. In our implementation, we have already incorporated a degree of heterogeneous computing by offloading output data from the GPU to the CPU for statistical computation. Such optimisations could be valuable in domains where rapid evaluation of complex architectures is necessary. 

Experimental results demonstrate that \texttt{epsinas} applies to convolutional networks and generalizes well to recurrent networks—with careful consideration for embedding layer initialization. The method can also be readily incorporated into other NAS frameworks, such as random search and evolutionary algorithms.

The main downside of our metric is that the optimal weight values need to be set separately for different search spaces and neural topologies. We found that using both high and low weight values helps highlight differences in initialization, leading to better correlations with final performance. While the weights should cover both extremes to test the architecture's robustness, it's unclear how to automate this selection process.

These results offer an intriguing clue about the inner workings of neural networks. We hope that \texttt{epsinas} 's performance can be improved with a better theoretical understanding of the optimal weight setup. We believe that further development of this method will lead to a faster and computationally efficient NAS and a clearer understanding of network prediction and generalization mechanisms.

These results might offer some interesting clues about how neural networks work. We are hopeful that with a better theoretical understanding of the optimal weight setup, \texttt{epsinas} 's performance can improve, and we believe that further development of this method could help us better understand network prediction and generalization.

\backmatter

\section*{Declarations}

\section*{Competing interests}
The authors have no competing interests to declare that are relevant to the content of this article.

\section*{Funding}
No funding was received for conducting this study.

\section*{Code availability}
The code  supporting this study can be found on GitHub at \url{https://github.com/egracheva/epsinas}.

\bibliography{main}% common bib file
%% if required, the content of .bbl file can be included here once bbl is generated
%%\input sn-article.bbl

\appendix

\section{Visualisation of other zero-cost NAS metrics}
\subsection{When integrated with other NAS methods}
\begin{figure*}[!t]
    \centering
    \begin{minipage}[l]{.34\textwidth}
        \subfigure{
        \includegraphics[width=.95\textwidth]{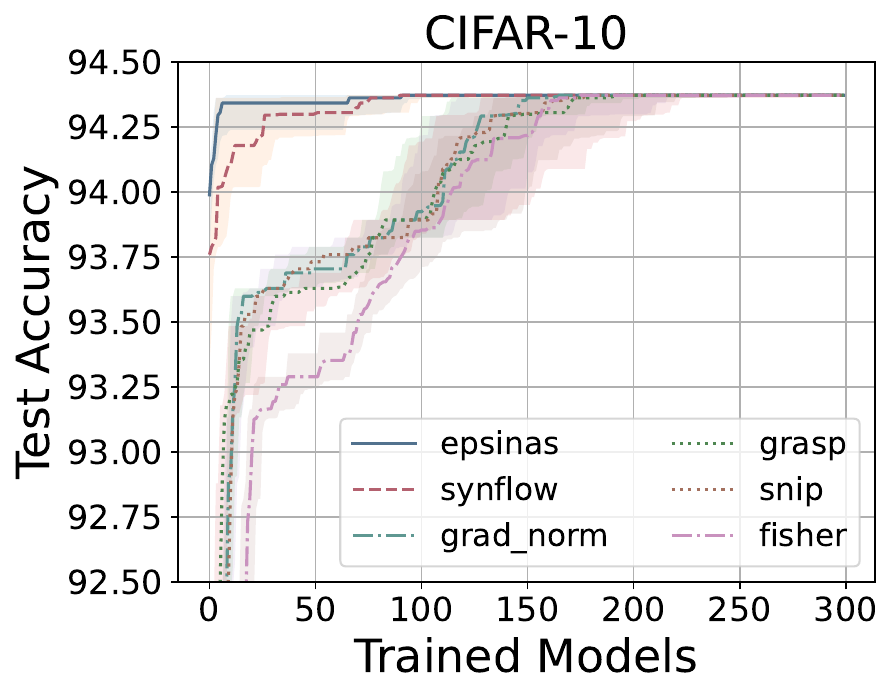}
        }\\
        \hspace{\fill}
        \subfigure{
        \includegraphics[width=.95\textwidth]{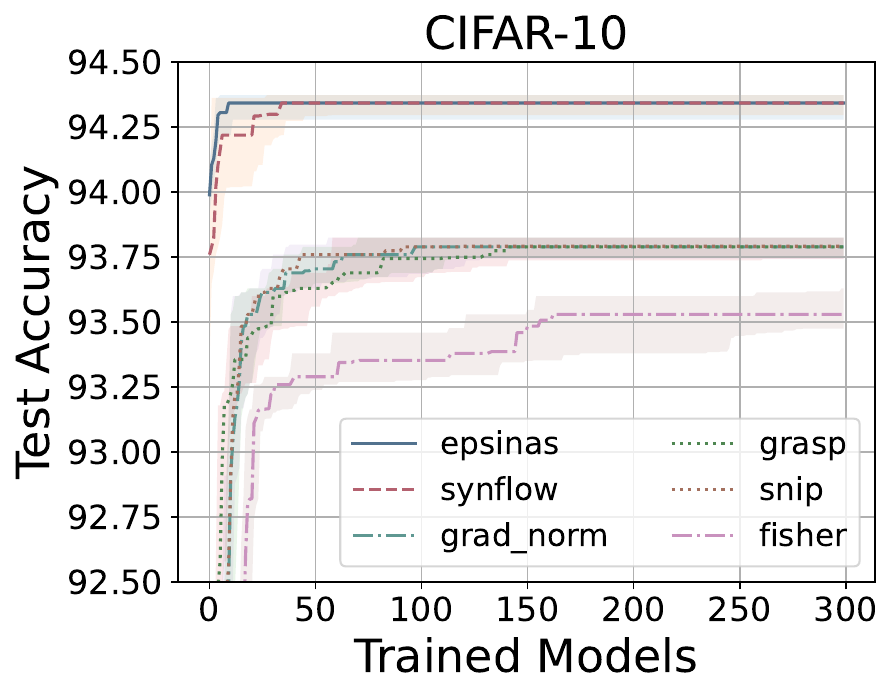}
        } 
    \end{minipage}
    \begin{minipage}[l]{.3\textwidth}
        \subfigure{
        \includegraphics[width=.95\textwidth]{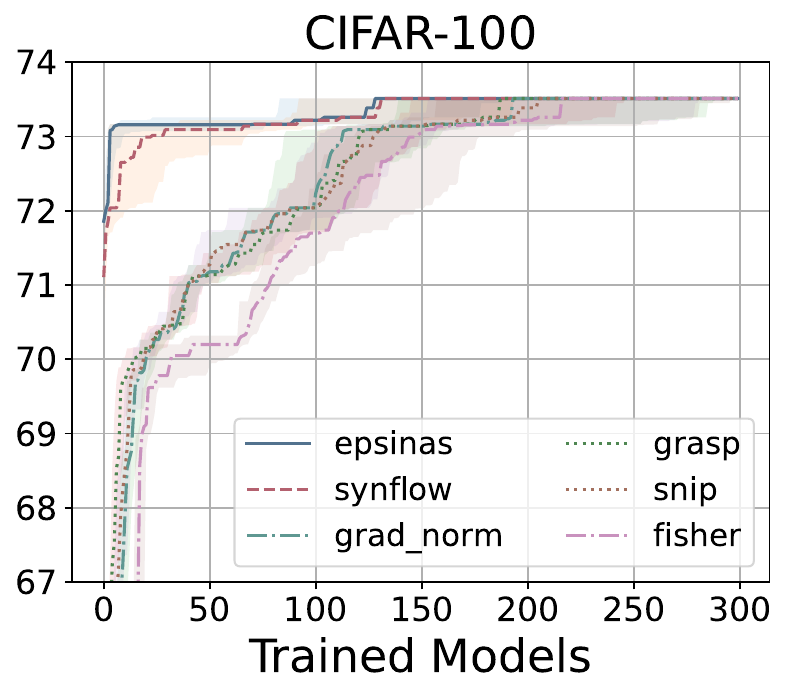}
        }\\
        \hspace{\fill}
        \subfigure{
        \includegraphics[width=.95\textwidth]{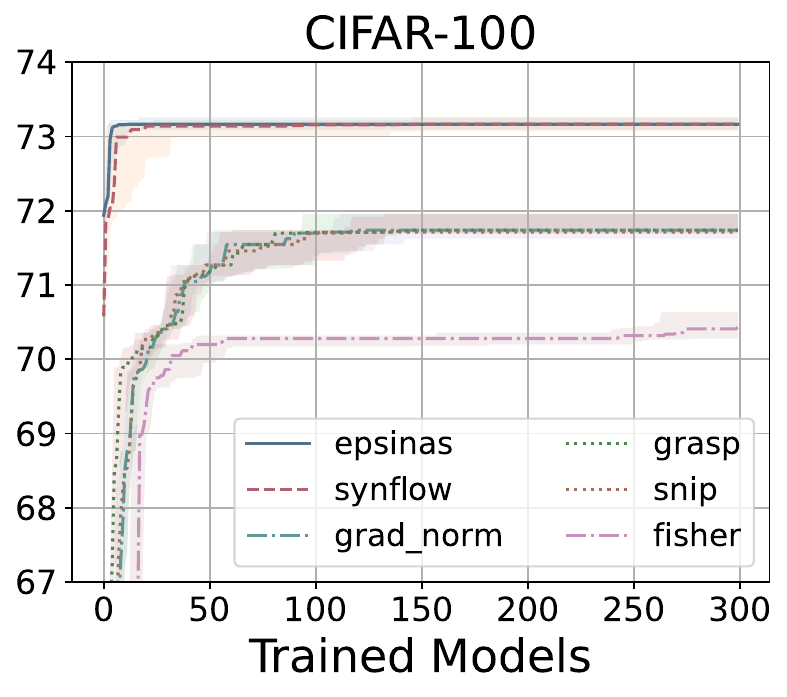}
        }
    \end{minipage}
    \begin{minipage}[l]{.3\textwidth}
        \subfigure{
        \includegraphics[width=.95\textwidth]{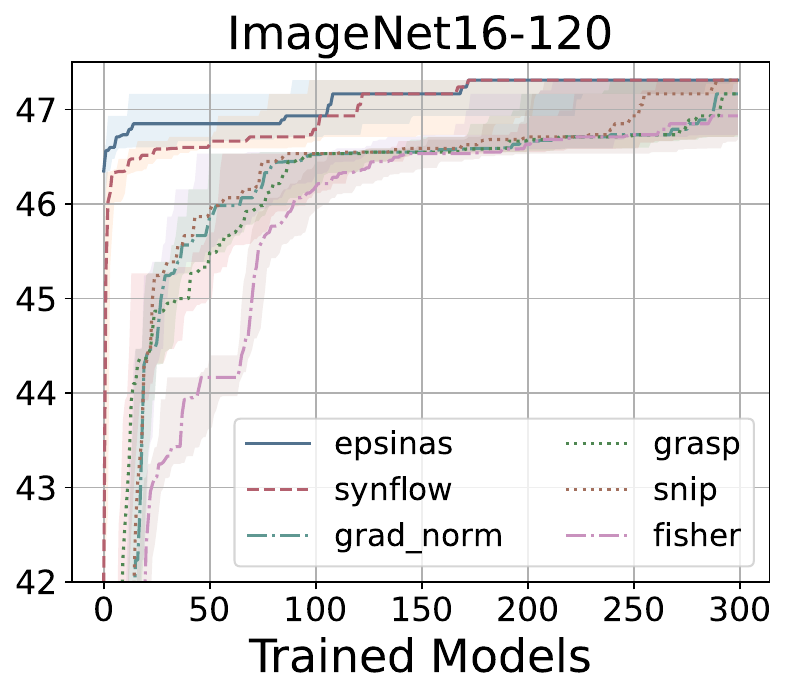}
        }\\
        \hspace{\fill}
        \subfigure{
        \includegraphics[width=.95\textwidth]{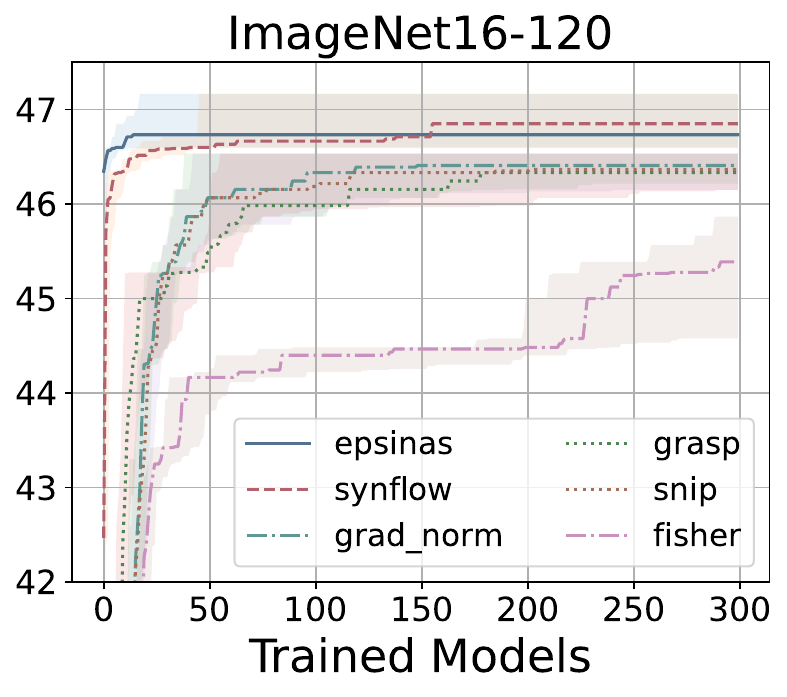}
        }
    \end{minipage}
\caption{Various zero-cost NAS metrics' performance when used as a warm-up for ageing evolution (top) and random search (bottom) for three datasets from the NAS-Bench-201 search space. Shadows span between lower and upper quartiles based on $100$ rounds. The warm-up population is $3000$ architectures.} 
\label{fig:integration_other}
\end{figure*}

In Section \ref{sec:implement}, we show how \texttt{epsinas} metric improves the performance of ageing evolution and random search when used for warming up. Figure \ref{fig:integration_other} compares \texttt{epsinas} integration to other metrics from \cite{abdelfattah2021zero}. For both EA and RS, we use the metrics for a warm-up and run the procedure until the number of trained architectures reaches $300$ with $100$ random rounds. The warm-up pool contains $3000$ randomly selected architectures.

\subsection{Correlation with accuracy}
In \cite{abdelfattah2021zero}, the metrics are presented through statistical measures, but we feel that visualisation helps to improve understanding. Here, we provide visualisations for two search spaces built on the data provided by the authors.

\begin{figure*}[!t]
\centering
\begin{minipage}{.9\textwidth}
\resizebox{.9\textwidth}{!}{%
    \includegraphics[height=3cm]{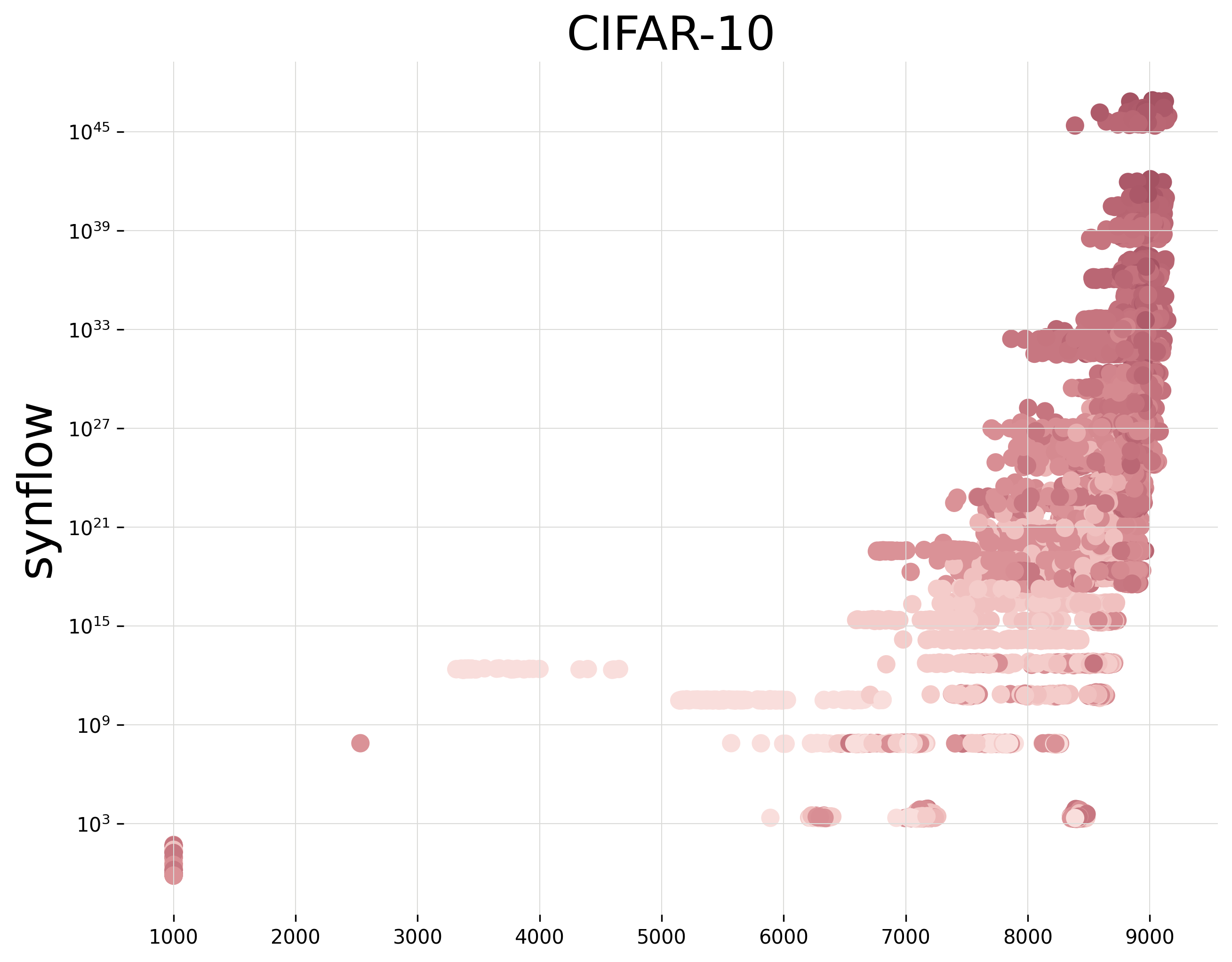}\hfill
    \includegraphics[height=3cm]{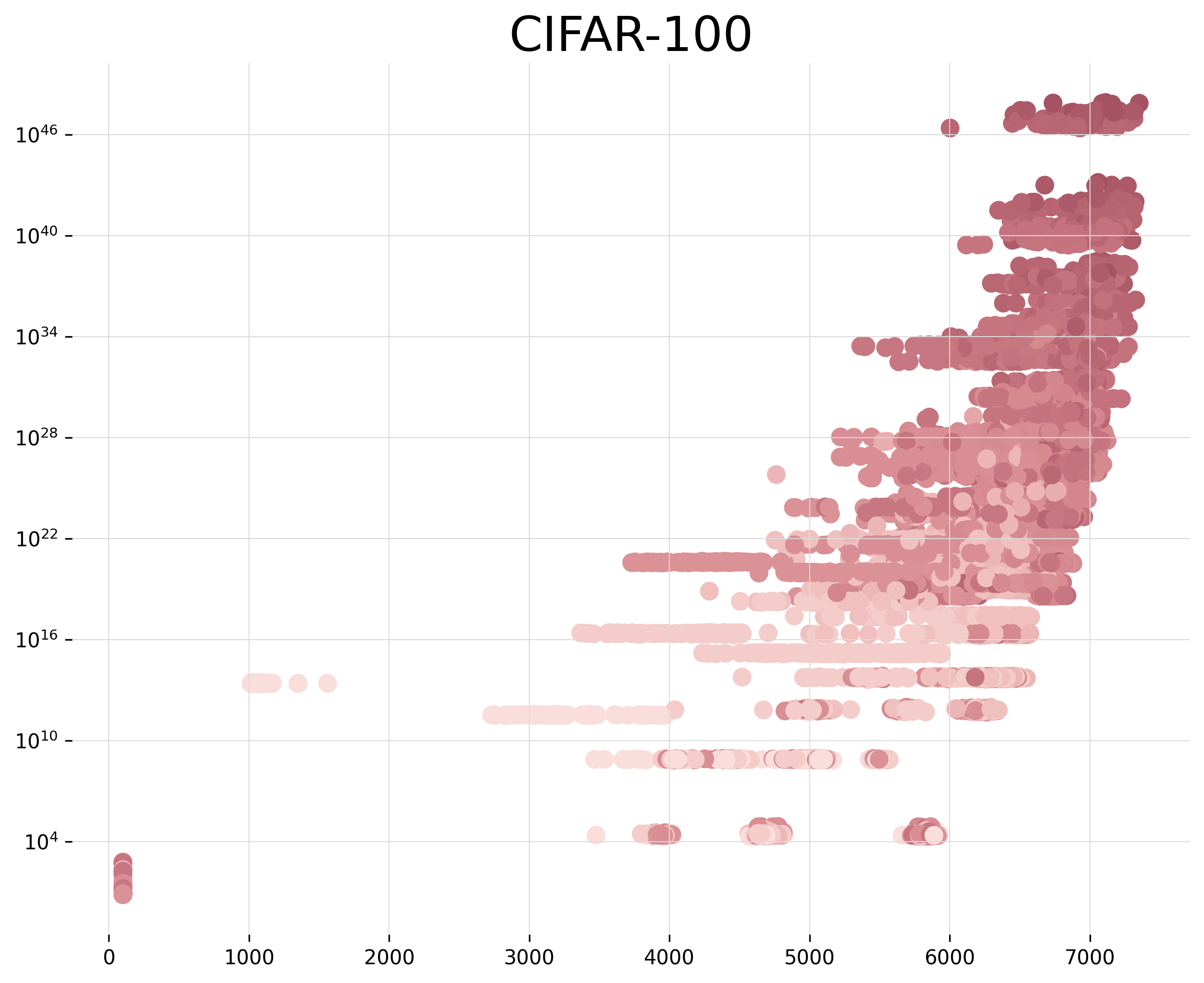}\hfill
    \includegraphics[height=3cm]{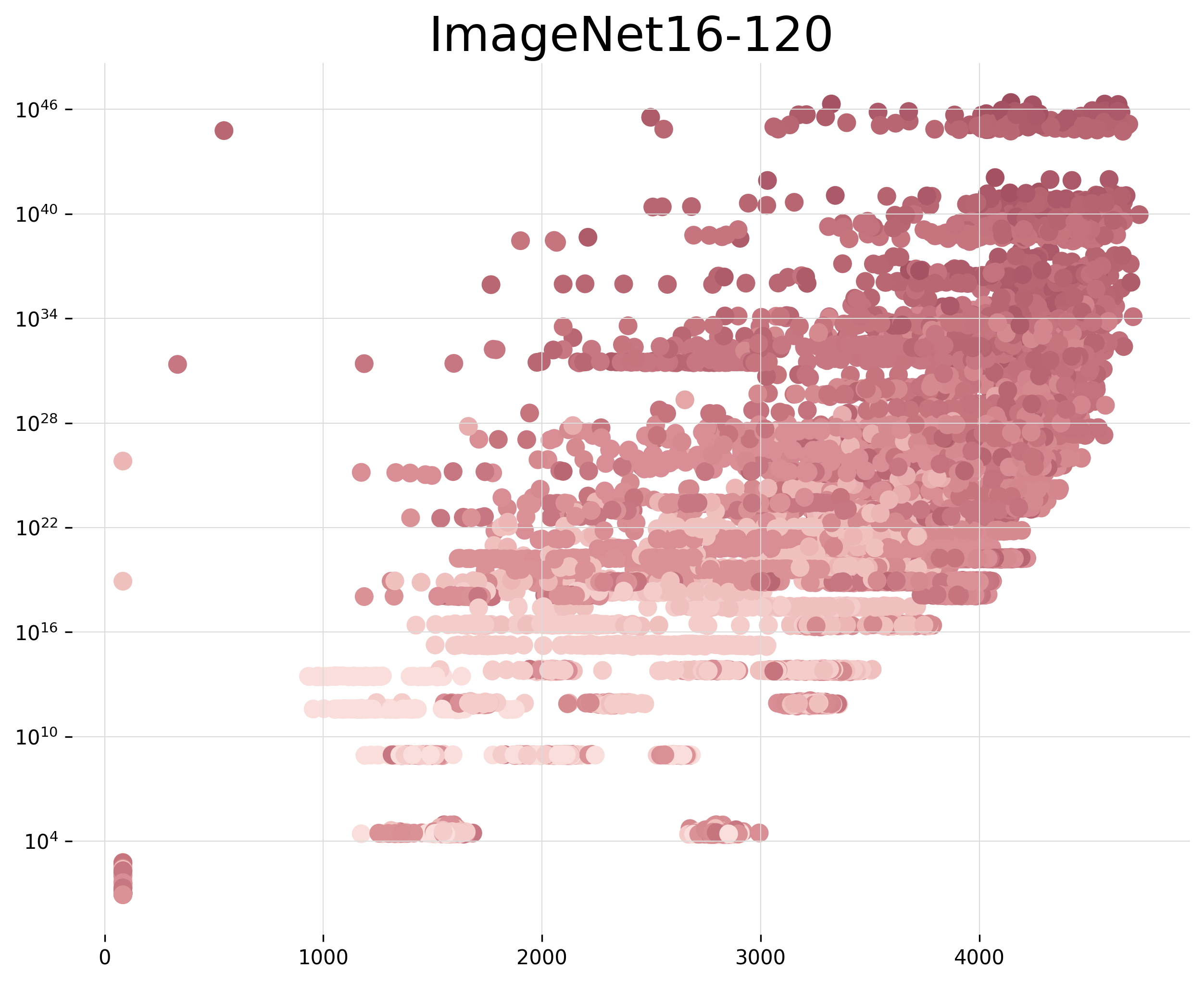}
}
\end{minipage}\hfill
\begin{minipage}{.9\textwidth}
\resizebox{.9\textwidth}{!}{%
    \includegraphics[height=3cm]{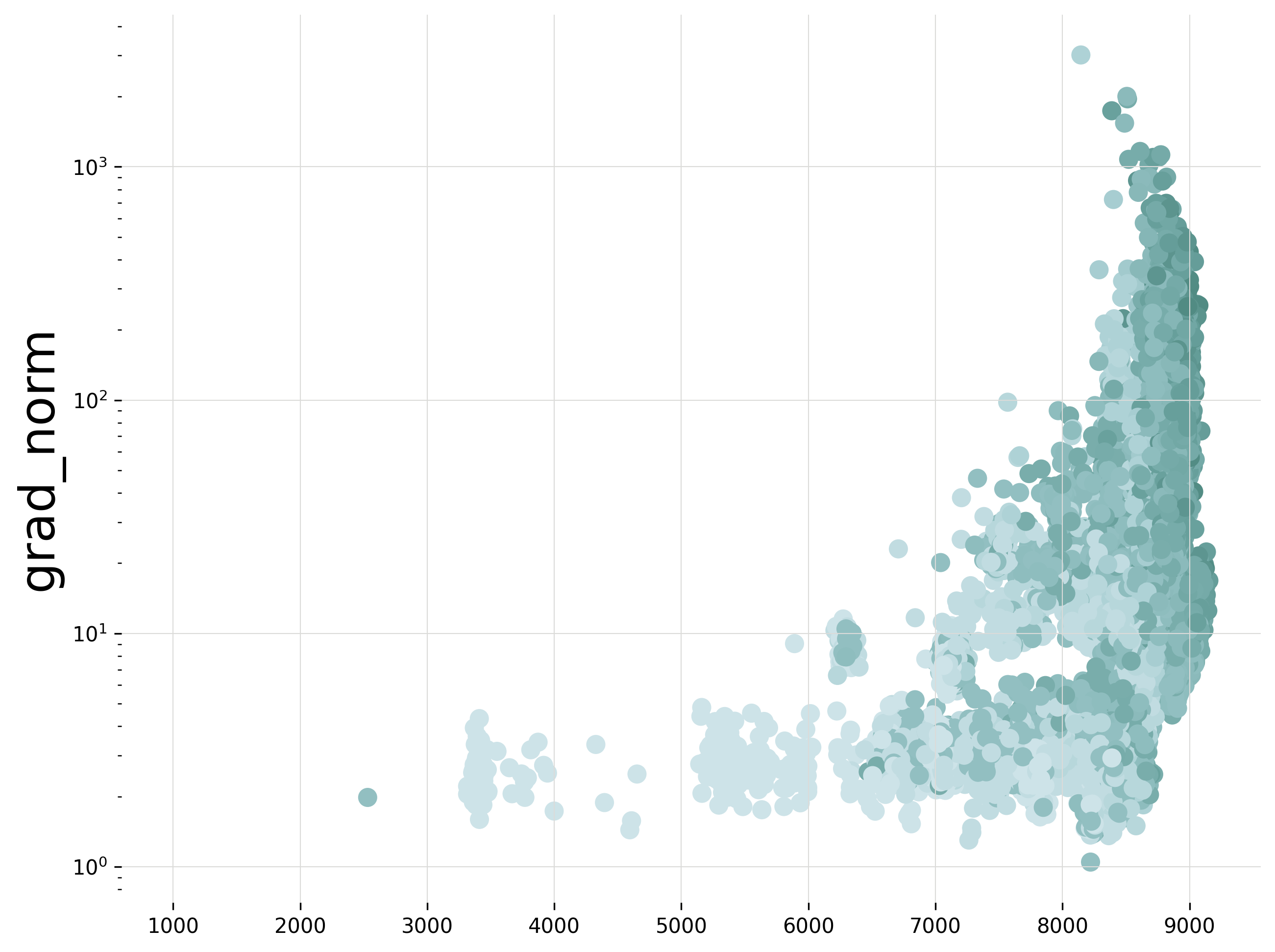}\hfill
    \includegraphics[height=3cm]{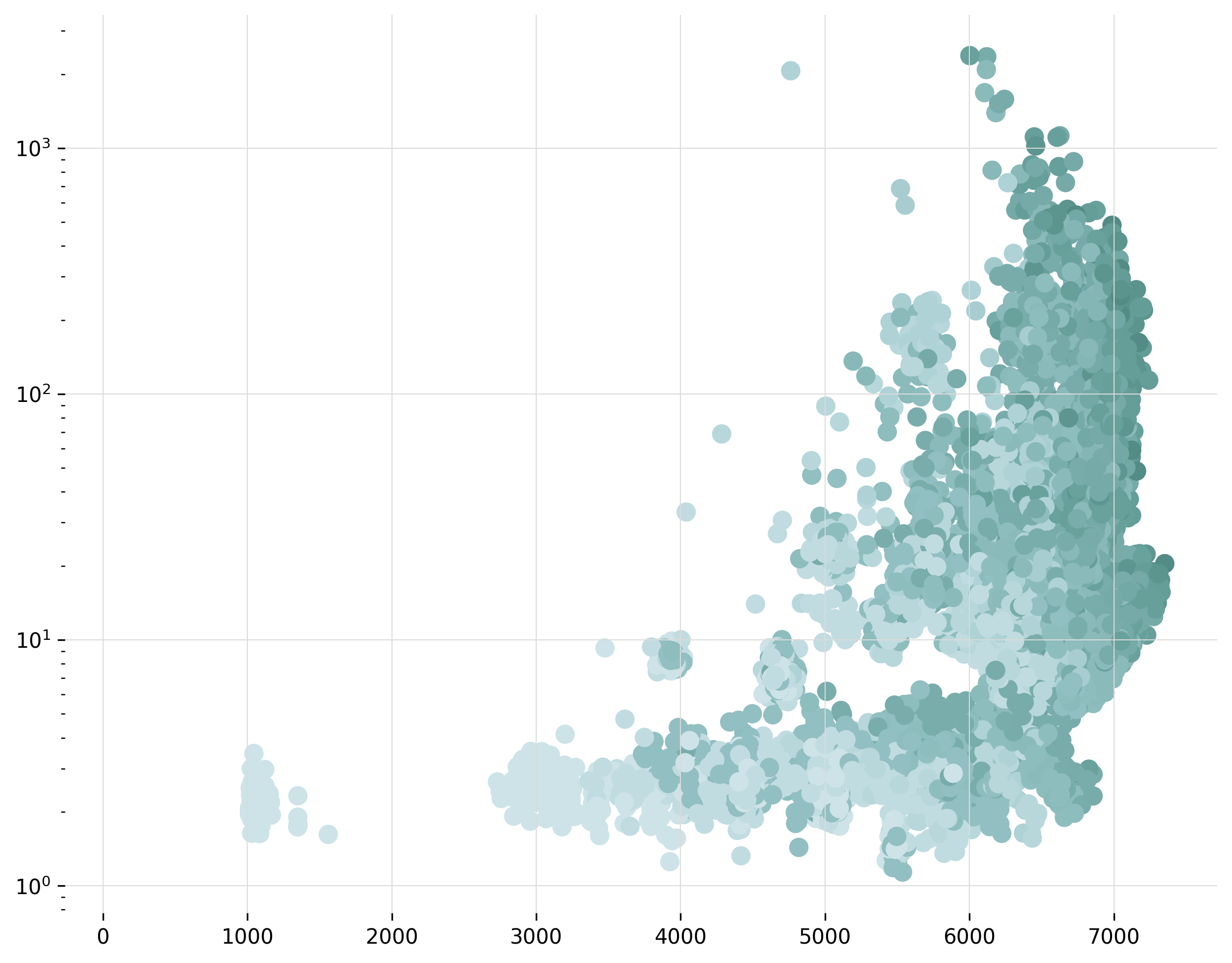}\hfill
    \includegraphics[height=3cm]{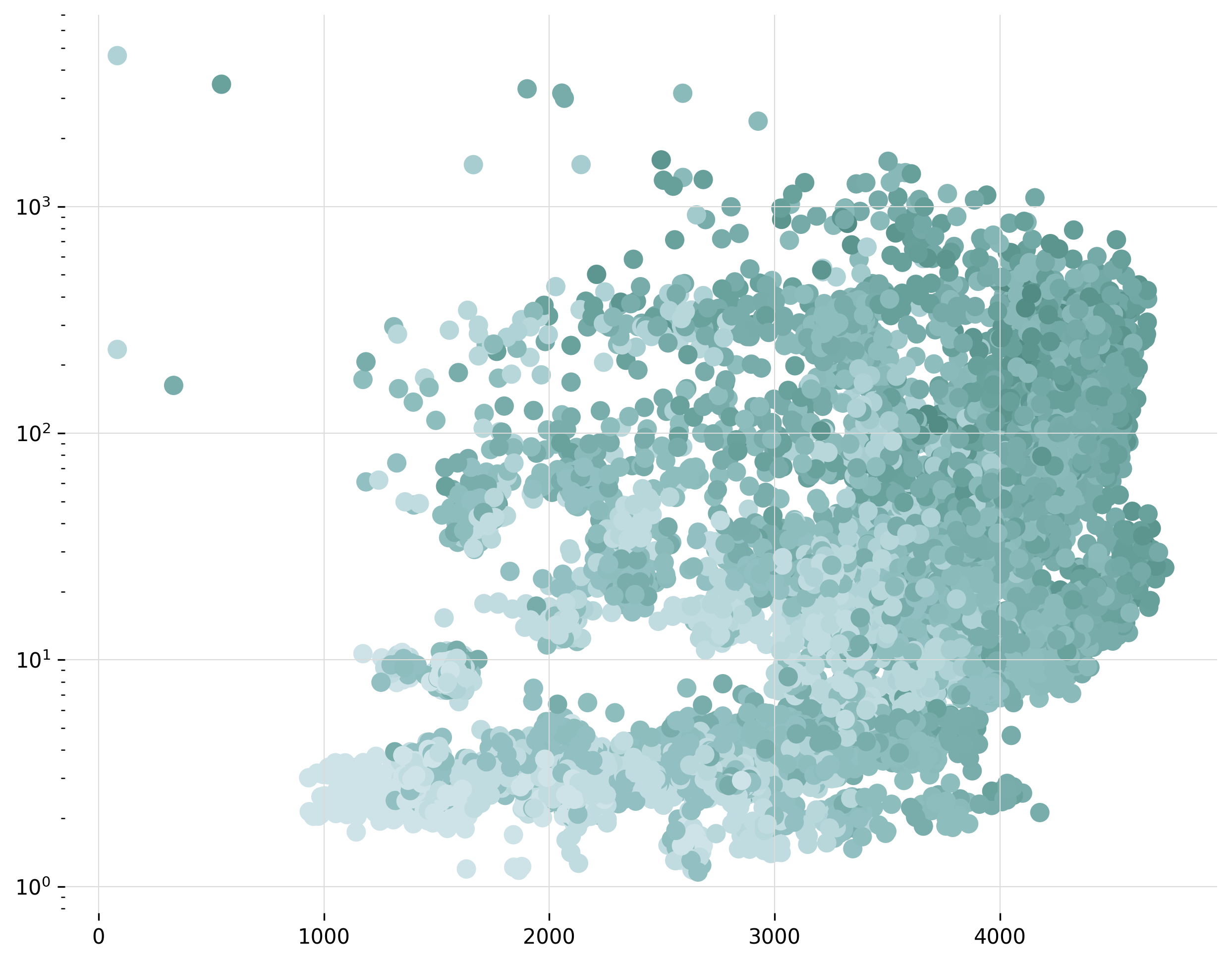}
    }
\end{minipage}\hfill
\begin{minipage}{.9\textwidth}
\resizebox{.9\textwidth}{!}{%
    \includegraphics[height=3cm]{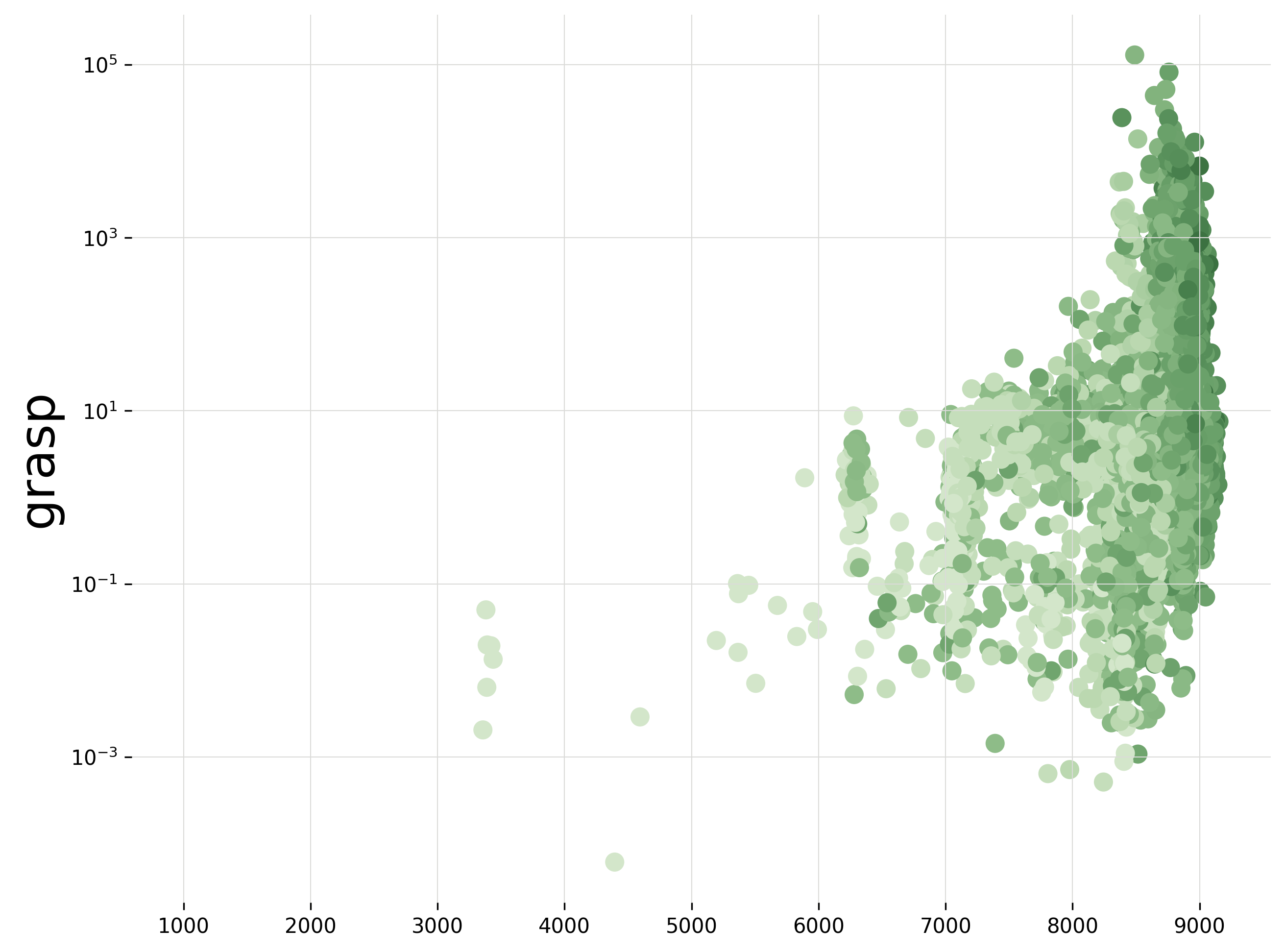}\hfill
    \includegraphics[height=3cm]{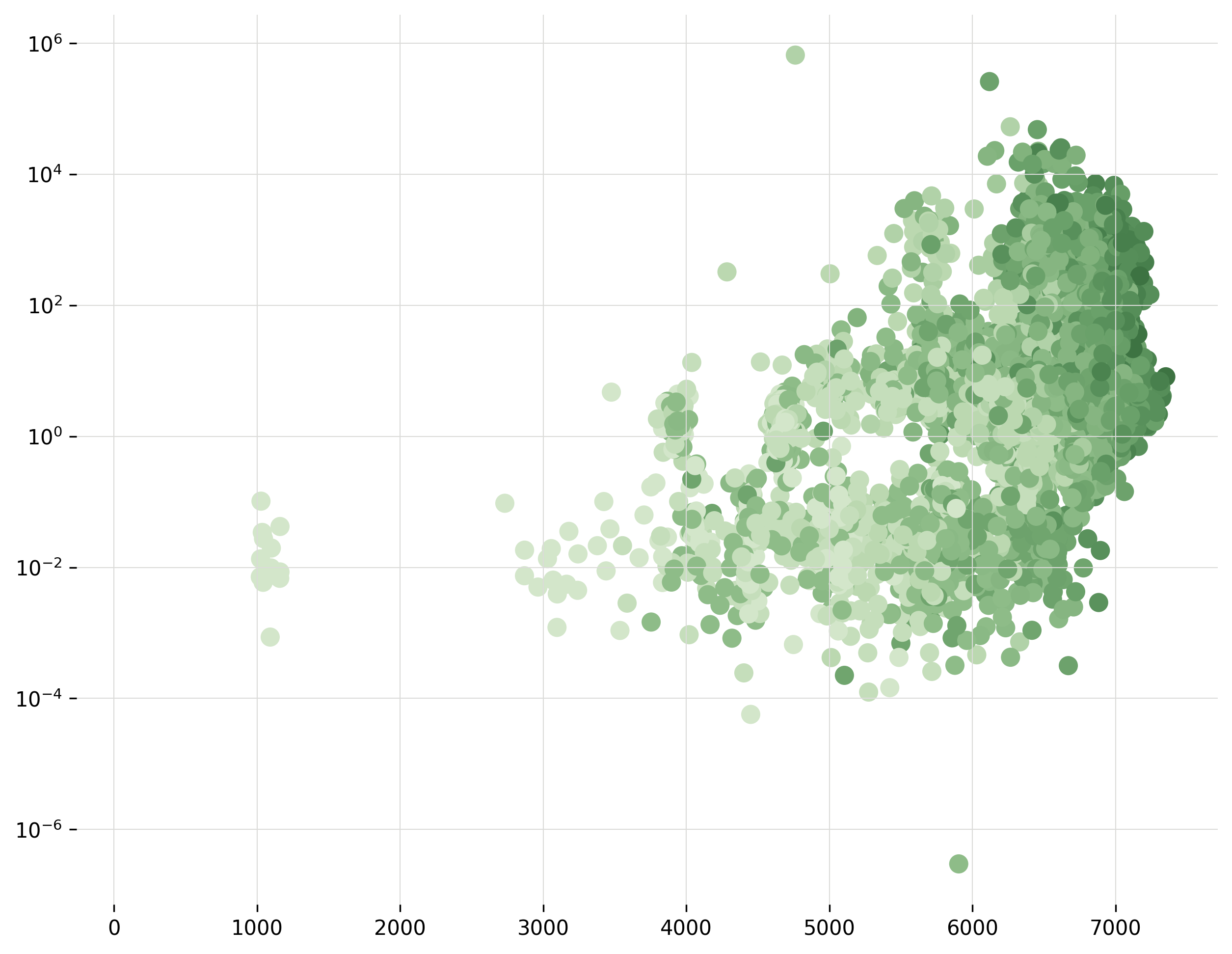}\hfill
    \includegraphics[height=3cm]{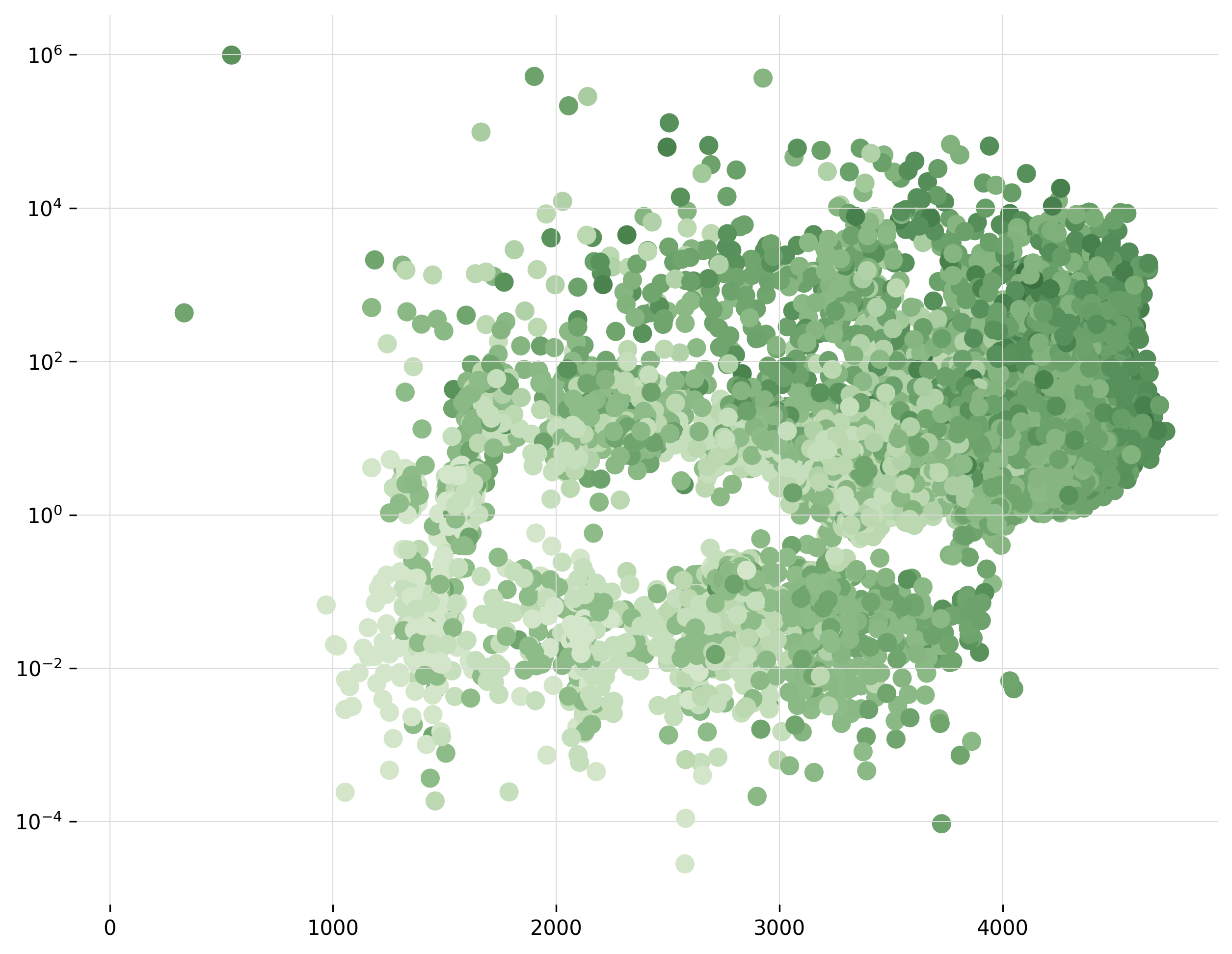}
    }
\end{minipage}\hfill
\begin{minipage}{.9\textwidth}
\resizebox{.9\textwidth}{!}{%
    \includegraphics[height=3cm]{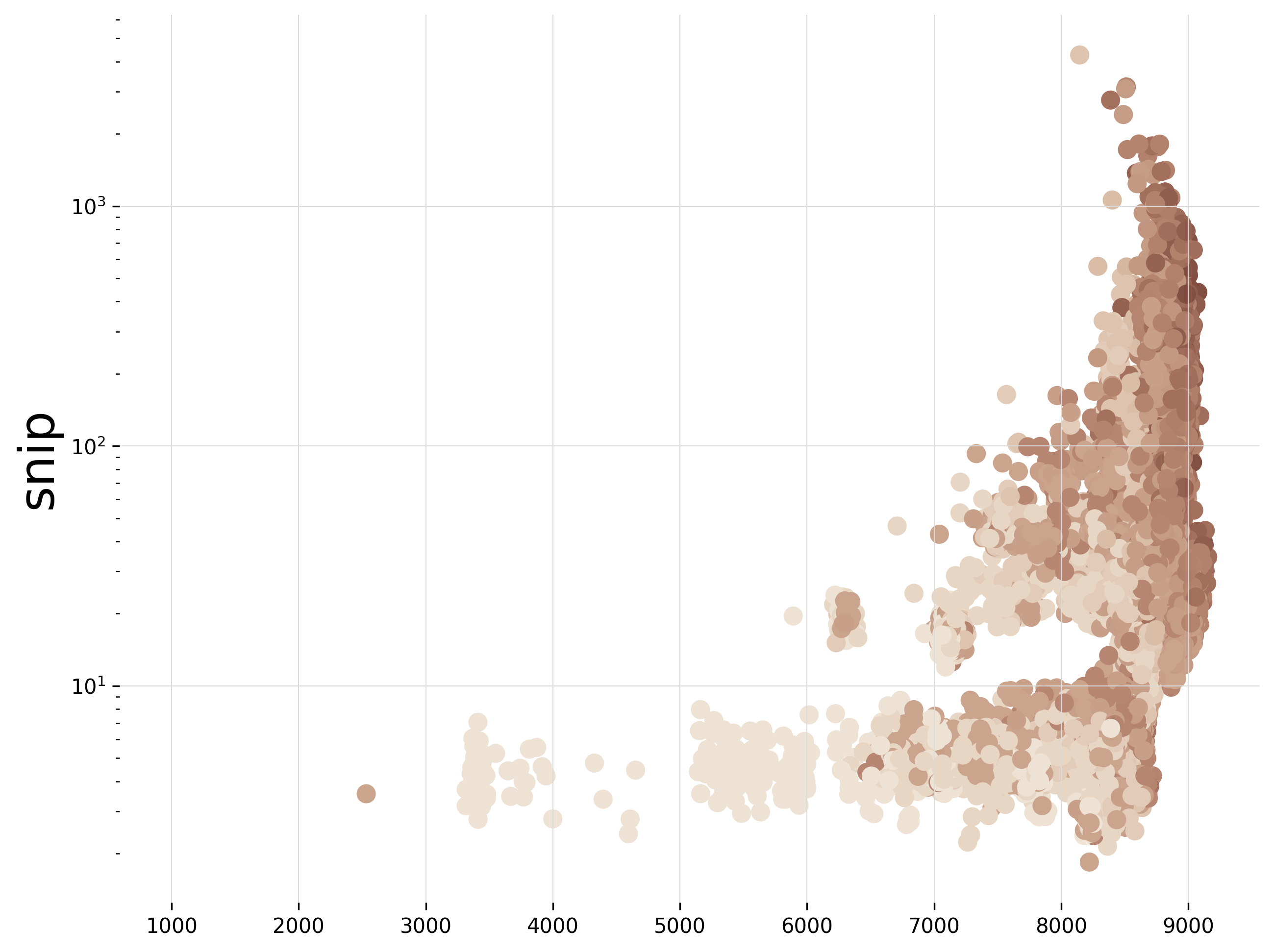}\hfill
    \includegraphics[height=3cm]{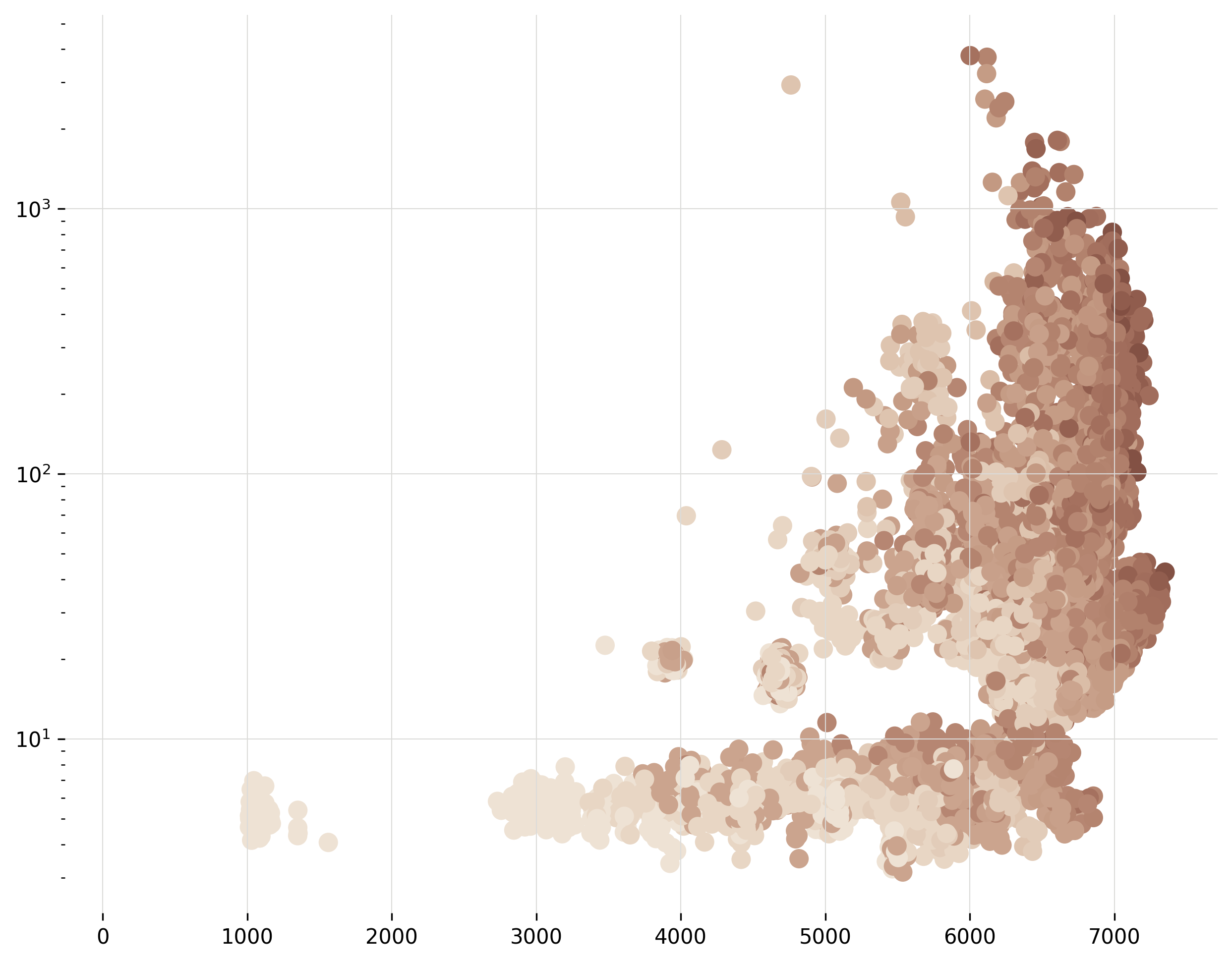}\hfill
    \includegraphics[height=3cm]{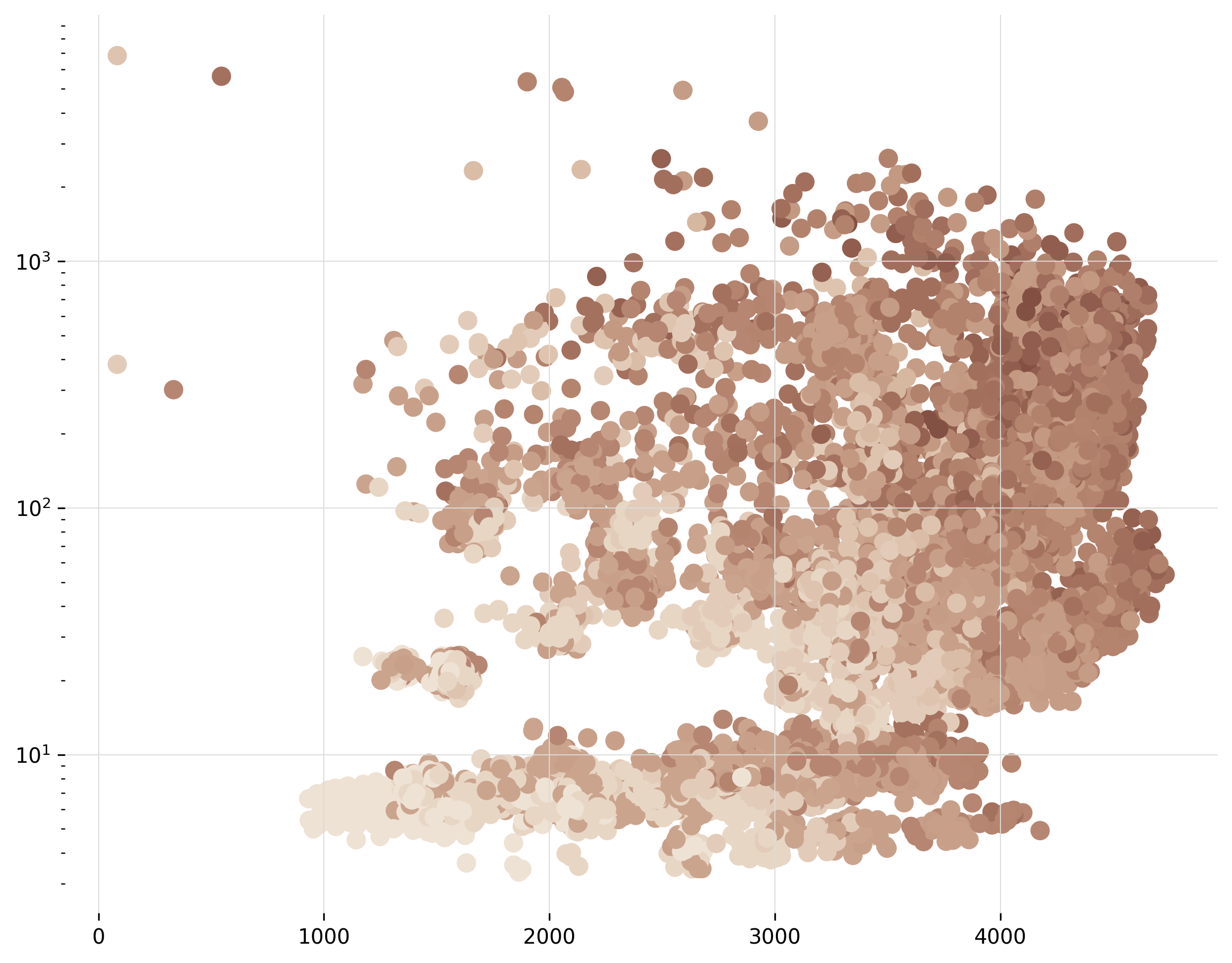}
    }
\end{minipage}\hfill
\begin{minipage}{.9\textwidth}
\resizebox{.9\textwidth}{!}{%
    \includegraphics[height=3cm]{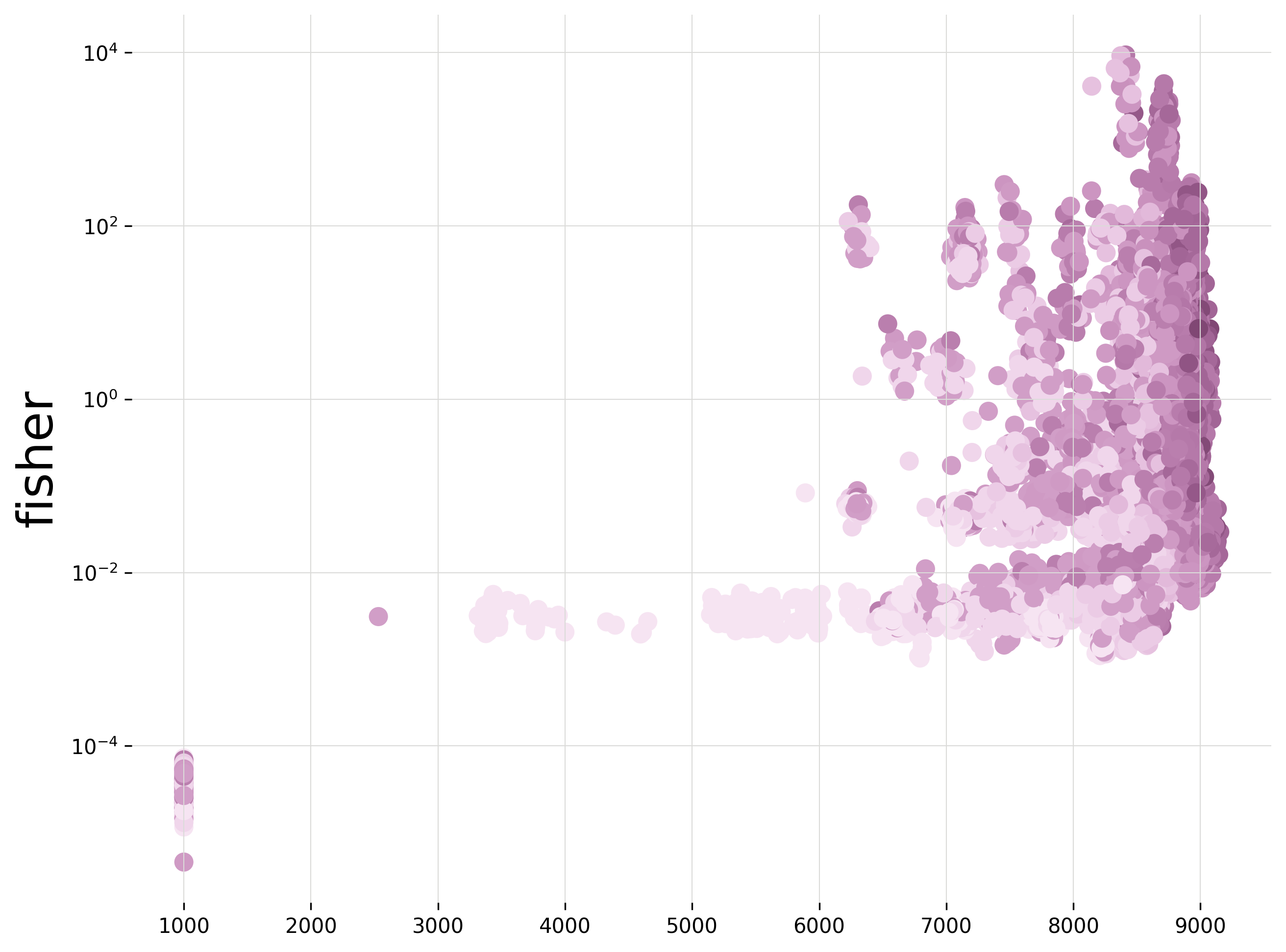}\hfill
    \includegraphics[height=3cm]{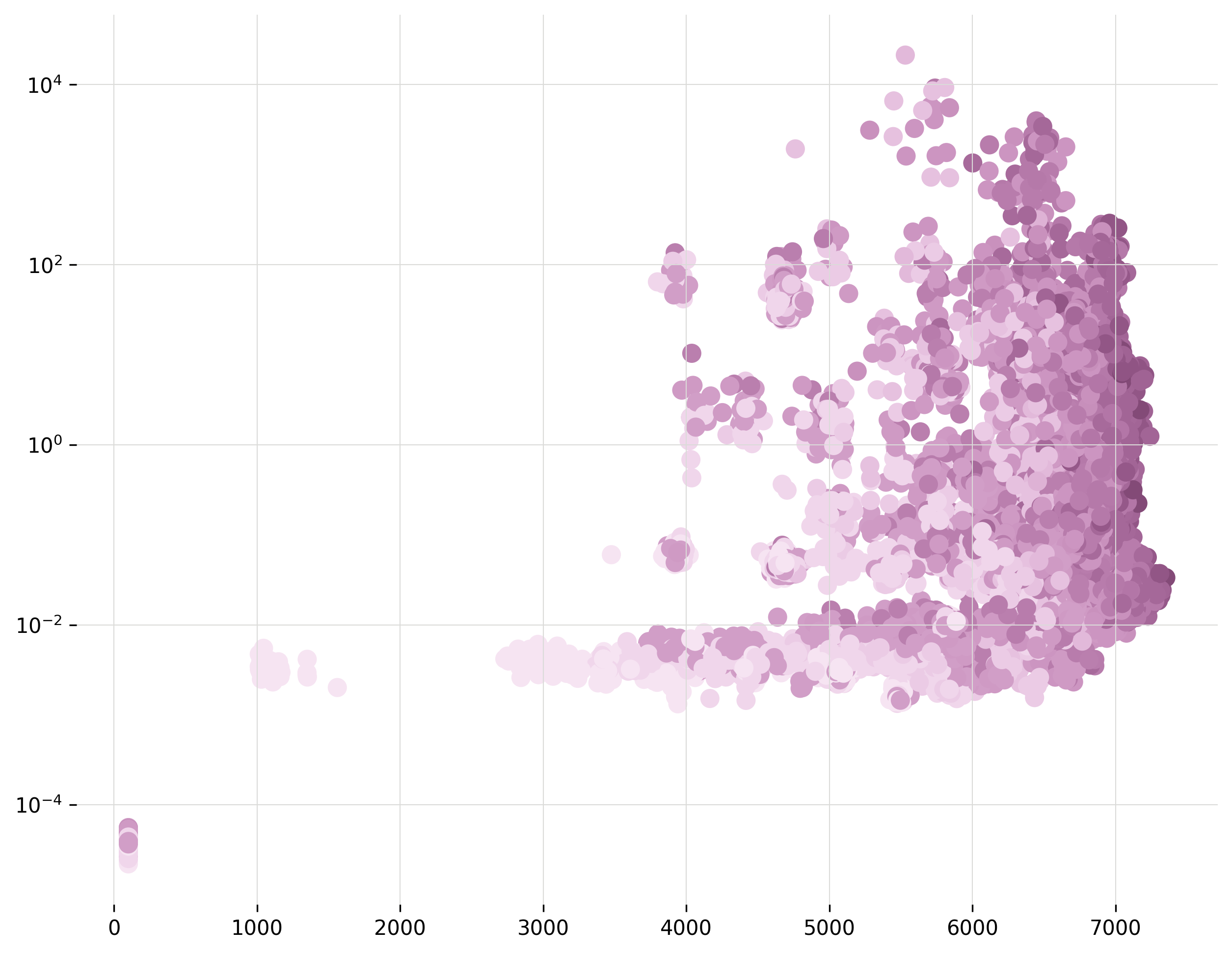}\hfill
    \includegraphics[height=3cm]{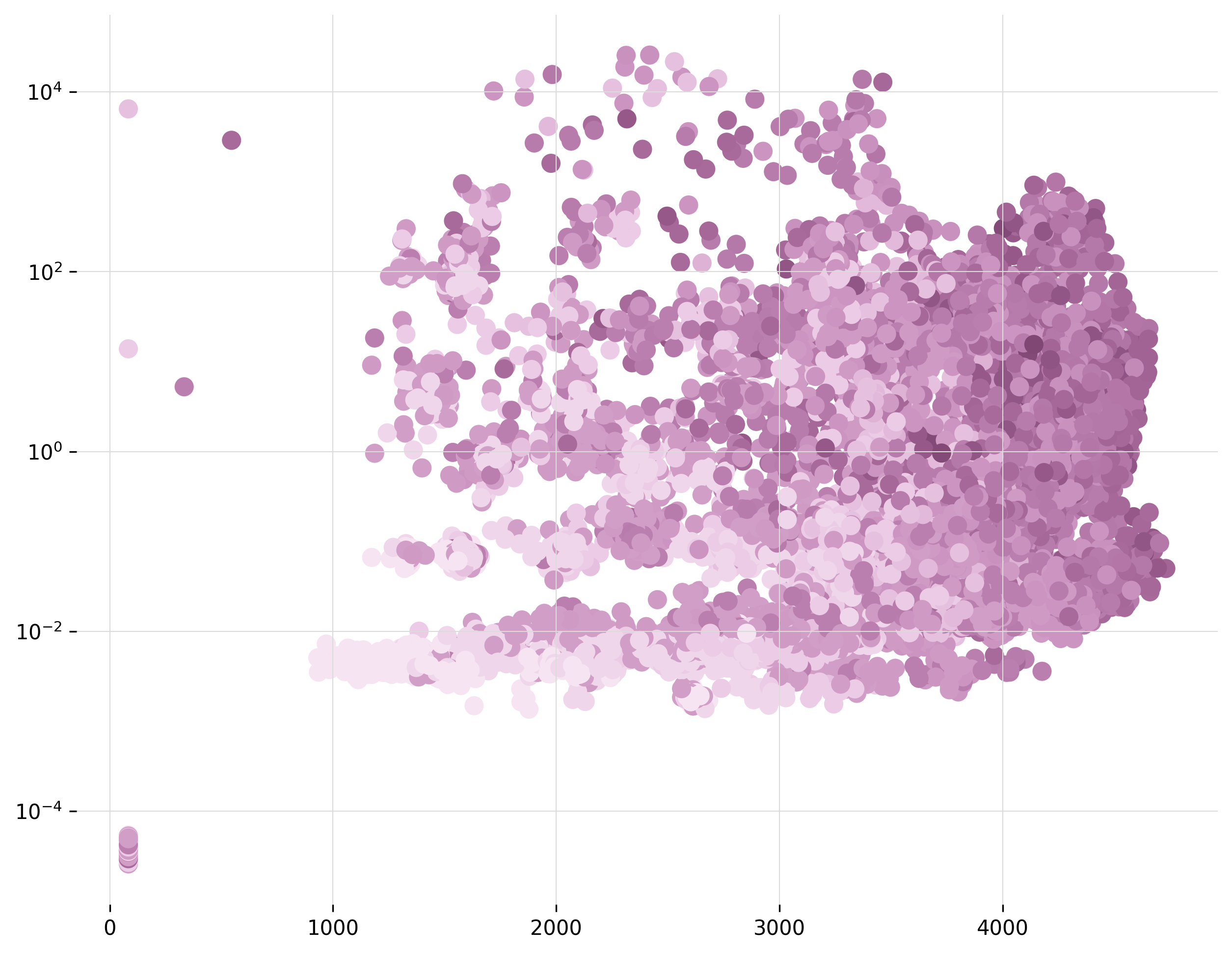}
    }
\end{minipage}
\caption{Zero-cost metrics performance illustration for NAS-Bench-201 search space evaluated on CIFAR-10, CIFAR-100 and ImageNet16-120 datasets, based on data from \cite{abdelfattah2021zero}. Each dot corresponds to an architecture; the darker the colour, the more parameters it contains. The figure represents the search space of $15{,}625$ networks (excluding architectures with \texttt{NaN} scores).}
\label{fig:other_results_201}
\end{figure*}

\end{document}